\documentclass[letterpaper, 10 pt, conference]{ieeeconf}  %

\IEEEoverridecommandlockouts                              %

\usepackage{cite}
\usepackage{amsmath,amssymb,amsfonts}
\usepackage{graphicx}
\usepackage{xcolor}
\usepackage[caption=false,font=footnotesize]{subfig}
\usepackage{booktabs}
\usepackage{multirow}
\usepackage{float}
\usepackage{hyperref}
\usepackage{tikz}
\usepackage{pgfplots}
\pgfplotsset{compat=1.18}
\usetikzlibrary{shapes.geometric}
\usetikzlibrary{patterns}
\usetikzlibrary{calc}

\definecolor{colorPP}{RGB}{255,214,31}
\definecolor{colorPRM}{RGB}{226,22,22}
\definecolor{colorRRT}{RGB}{160,28,187}
\definecolor{colorConservative}{RGB}{255,127,14}
\definecolor{colorReactive}{RGB}{44,160,44}
\definecolor{colorConservativeSkill}{RGB}{255,165,0}
\definecolor{colorConservativeCtrl}{RGB}{230,92,0}
\definecolor{colorReactiveSkill}{RGB}{79,207,79}
\definecolor{colorReactiveCtrl}{RGB}{31,112,31}
\definecolor{colorStartGoal}{RGB}{96,96,96}
\definecolor{colorSkillFill}{RGB}{105,105,105}
\definecolor{colorPhasePurple}{RGB}{220,207,239}
\definecolor{colorPhaseBlue}{RGB}{205,231,240}
\definecolor{colorPhasePink}{RGB}{255,211,217}
\definecolor{colorPhaseGray}{RGB}{244,244,244}
\definecolor{colorBoxBorder}{RGB}{102,102,102}
\definecolor{colorTerminalBorder}{RGB}{153,153,153}

\usepackage[capitalise]{cleveref}
\usepackage{subcaption}

\usepackage[tracking=true]{microtype}

\DeclareMathOperator*{\argmin}{arg\,min}

\newcommand{\colorsq}[1]{\textcolor{#1}{\rule[0.12ex]{1.1ex}{1.1ex}}}

\title{\LARGE \bf
Multi-Robot Multi-Goal Motion Planning with Stochastic Skills
}

\author{William Schnyder, Valentin N. Hartmann, Stelian Coros
\thanks{All authors are with the Computational Robotics Lab (CRL), ETH Zürich.}
}

\begin{document}

\maketitle
\thispagestyle{empty}
\pagestyle{empty}

\begin{abstract}
\label{sec:abstract}
As robots are increasingly deployed in groups and share workspaces to execute real-world tasks, planning their concurrent motions around complex manipulation skills becomes essential.  
These skills involve continuous physical execution and may exhibit stochastic behavior, resulting in variable execution times and uncertain continuous trajectories. 
Existing planners either limit execution to single-robot scenarios, rely on open-loop paths, or use post-hoc scheduling that prevents dynamic coordination. 
In this paper, we address this gap by integrating stochastic skills into sampling-based multi-robot planning by formulating the problem as a Markov Decision Process (MDP) over a multi-modal composite roadmap. 
For stochastic skills, solving the MDP yields a reactive policy that allows controllable robots to dynamically adapt their motions in response to other robots' execution of manipulation skills. 
By resolving skill uncertainty directly at planning time, this approach avoids the pessimism of conservative baselines and unlocks robust, dynamic multi-robot coordination. 
Code for the planners is available at \href{https://www.vhartmann.com/stochastic-skills}{\textls[-40]{\texttt{www.vhartmann.com/stochastic-skills}}}.
\end{abstract}

\section{Introduction}
\label{sec:intro}
In modern industrial and automated environments, maximizing throughput is a central objective. This naturally drives the adoption of multi-robot systems, where several robots share a single workspace to execute tasks in parallel rather than in sequence. 
Realizing this gain requires reasoning jointly about the motion of all robots and the tasks they must complete, since robots operating in the same space can block, delay, or collide with one another. 
Relying on decoupled approaches, such as prioritized planning, avoids the complexity of the composite state space but often recovers only a fraction of the achievable throughput.

While motion planners excel at collision-free navigation, real-world tasks like grasping involve complex physical contact dynamics. 
To abstract this, robots typically execute tasks as \emph{skills} driven by feedback controllers or learned policies, rather than following purely geometric paths. 
Integrating these skills into a multi-robot planner presents two major challenges. 
First, a skill involves a continuous physical execution governed by an autonomous low-level controller that the motion planner cannot steer.
Consequently, while the skill-executing robot acts as an uncontrollable dynamic agent during skill execution, the planner actively coordinates the remaining \emph{controllable} robots to operate safely in the shared workspace.
Second, physical skills are inherently \emph{stochastic}. 
In real-world assembly tasks, uncertainty requires the controller to adjust, leading to variable execution times (e.g., a fast grasp vs. a slower one) or spatial trajectory variations. 
Rather than ignoring this uncertainty until execution, our planner models skills as stochastic distributions.
As a result, a single deterministic path is no longer sufficient. 
Coordinating concurrent motion around these variable outcomes requires computing a closed-loop \emph{policy} specifying how surrounding robots should dynamically react.

\begin{figure}[t]
    \centering
    \includegraphics[width=\columnwidth]{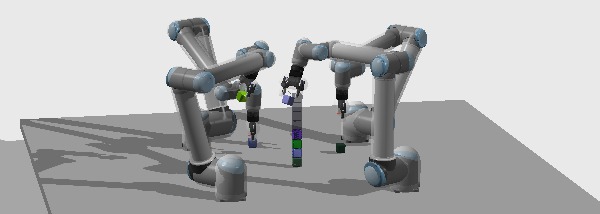}\\
    \vspace{5pt}
    \centerline{\begin{tikzpicture}
\begin{axis} [
  width=\linewidth,
  height=0.5\textwidth,
  unbounded coords=jump,
  xtick align=inside,
  ytick align=inside,
  anchor=north,
  hide axis,
  clip=false,
  xmajorgrids,
  ymajorgrids,
  major grid style={densely dotted, black!20},
  xmin=0,
  xmax=10,
  ymin=0,
  ymax=10,
  xlabel style={font=\footnotesize},
  xticklabel style={font=\footnotesize},
  ylabel style={font=\footnotesize},
  yticklabel style={font=\footnotesize},
  legend style={anchor=south, legend cell align=left, legend columns=6, at={(axis cs:5, 6)}, font=\small}
]

\addlegendimage{legend image code/.code={
  \fill[colorSkillFill] (0cm,-0.08cm) rectangle (0.16cm,0.08cm);
  \draw[red, line width=0.9pt] (0cm,-0.08cm) rectangle (0.16cm,0.08cm);
}}
\addlegendentry{Skill (Pick)}

\addlegendimage{legend image code/.code={
  \fill[colorSkillFill] (0cm,-0.08cm) rectangle (0.16cm,0.08cm);
  \draw[colorBoxBorder, line width=0.5pt, pattern=north east lines, pattern color=black] (0cm,-0.08cm) rectangle (0.16cm,0.08cm);
}}
\addlegendentry{Wait}

\addlegendimage{legend image code/.code={
  \fill[colorPhasePurple] (0cm,-0.08cm) rectangle (0.16cm,0.08cm);
  \draw[colorBoxBorder, line width=0.5pt] (0cm,-0.08cm) rectangle (0.16cm,0.08cm);
}}
\addlegendentry{Pre Pick}

\addlegendimage{legend image code/.code={
  \fill[colorPhaseBlue] (0cm,-0.08cm) rectangle (0.16cm,0.08cm);
  \draw[colorBoxBorder, line width=0.5pt] (0cm,-0.08cm) rectangle (0.16cm,0.08cm);
}}
\addlegendentry{Place}

\addlegendimage{legend image code/.code={
  \fill[colorPhasePink] (0cm,-0.08cm) rectangle (0.16cm,0.08cm);
  \draw[colorBoxBorder, line width=0.5pt] (0cm,-0.08cm) rectangle (0.16cm,0.08cm);
}}
\addlegendentry{Home}

\addlegendimage{legend image code/.code={
  \fill[colorPhaseGray] (0cm,-0.08cm) rectangle (0.16cm,0.08cm);
  \draw[colorTerminalBorder, line width=0.5pt] (0cm,-0.08cm) rectangle (0.16cm,0.08cm);
}}
\addlegendentry{Terminal}

\end{axis}
\end{tikzpicture}}
    \vspace{5pt}
    \includegraphics[width=\columnwidth]{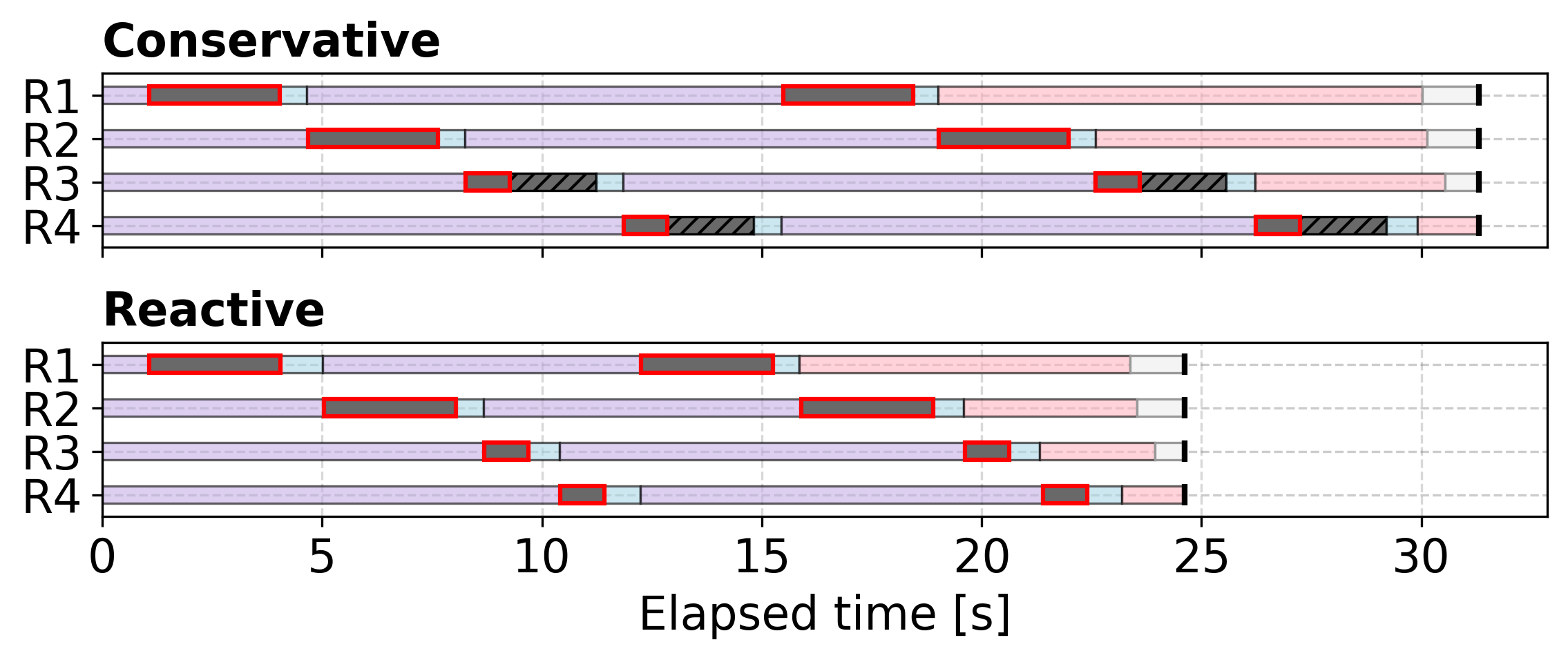}
    \caption{Four-robot execution of the eight-object stacking task with stochastic pick durations.
    Top: manipulation environment. 
    Bottom: execution schedules for the same skill realization. 
    The reactive policy adapts online to the realized pick durations, avoiding the idle time introduced by the conservative worst-case schedule.}
    \label{fig:3d_gantt}
\end{figure}

The multi-modal multi-robot multi-goal (MRMG) formulation of Hartmann~et~al.~\cite{Hartmann2026SamplingBased} provides composite-space planners for multi-robot problems.
However, it assumes purely geometric planning and cannot handle the uncertain execution of real-world skills.
To address this gap, we extend MRMG planning to integrate skills natively at planning time. 

Within this framework, we treat deterministic skill planning as a noise-free special case. 
As skill uncertainty increases, open-loop deterministic plans fail. 
To solve planning problems in stochastic environments, we formulate the problem as a Markov Decision Process (MDP) over a multi-modal composite-space roadmap, and compute a reactive policy that minimizes expected cost and allows controllable robots to adapt their motions in real time as skills execute.

We evaluate our framework across three experimental settings. 
First, we validate the native integration of skills into multi-robot planning under nominal, deterministic execution, comparing how different underlying planner architectures handle continuous skills. 
Second, in environments with stochastic skills, we compare our closed-loop reactive policy against a conservative open-loop baseline to highlight the advantages of real-time adaptation over worst-case scheduling. 
Finally, we evaluate our reactive planner on multi-arm manipulation tasks to demonstrate scalability to higher-dimensional composite state spaces.

\section{Related Work}
\label{sec:related}
\subsection{Planning with Manipulation Skills}
\label{sec:related:planning_skills}
A skill is a temporally extended, closed-loop behavior whose execution the planner does not steer. 
Such behaviors are commonly modeled as options, which assume that the planner only needs to make a decision at the temporal boundaries of the skill~\cite{Sutton1999Options}. 
In a multi-robot shared workspace, however, this abstraction fails: the planner must continuously coordinate controllable robots to avoid collisions with the autonomously executing skill.

Existing planners compose skills sequentially for a single robot. 
Task and Skill Planning (TASP) wraps skill policies in head and tail motion plans~\cite{Abbatematteo2024CIP} and chains them with a Task and Motion Planning (TAMP) solver, allowing force-controlled skills to be sequenced with kinematic motion~\cite{Hedegaard2026TaskSkillPlanning, Garrett2021TAMP}.
MOSAIC instead lets the skills guide the search via physics simulation, handling stochastic skills by summarizing a batch of rollouts into a success fraction that penalizes graph edge costs to yield a single nominal skill sequence~\cite{Mishani2026MOSAIC}.

For multiple robots, skills are typically coordinated only after planning. 
Multi-Robot Asynchronous Planning and Execution (APEX-MR) produces a turn-based sequential plan in which only one robot moves or executes a skill at a time, with each task solved as a single-robot motion planning problem, and skills entering only through a nominal reference path used for collision checking~\cite{Huang2025APEXMR}. 
APEX-MR then recovers parallelism and absorbs timing variations online by post-processing its sequential plan into a temporal plan graph (TPG)~\cite{Hoenig2016MAPFPOST}.

\subsection{Planning under Uncertainty}
\label{sec:related:planning_uncertainty}
Planning under uncertainty is commonly formulated as computing a policy over a discretization of the state space. 
The Stochastic Motion Roadmap (SMR) tackles this by randomly sampling configurations and simulating the robot's motion model to estimate transition probabilities. 
This transforms the roadmap into an MDP that can be solved with Dynamic Programming (DP) to maximize the probability of reaching the goal~\cite{Alterovitz2008SMR,Luna2014BMDP}.
Incremental algorithms, such as incremental Markov Decision Process (iMDP) and incremental Policy (iPolicy), refine this construction over time by expanding the state space and updating the policy iteratively, yielding anytime solutions that converge to the optimal one~\cite{Huynh2012IncrementalSampling, Zhao2024iPolicy}.

Other approaches make a single open-loop trajectory robust~\cite{Chang2026VIMP}, or plan in belief space under imperfect state observation~\cite{Aghamohammadi2014FIRM}.
While their observability assumptions vary, these methods primarily address environmental uncertainty such as dynamic obstacles, sensing noise, or the motion disturbances of a single controlled robot. 

The Anytime Task and Motion Markov Decision Process (ATM-MDP) computes branching task and motion policies for a single robot by coupling a stochastic shortest-path solver with a motion planner~\cite{Shah2020AnytimeIntegrated}.
To handle uncertainty, ATM-MDP branches on the discrete success or failure of high-level tasks, such as a grasp failing. 
In contrast, uncertainty in multi-robot settings has largely been handled by abstracting away continuous motion entirely, relying instead on discrete graphs to solve task allocation or robust scheduling against timing delays and task failures~\cite{Atzmon2020Robust, Atzmon2020Probabilistic, Choudhury2020SCoBA}.

Prior work either addresses continuous uncertainty for a single robot or coordinates multi-robot skills through open-loop and post-hoc methods. 
Bridging these domains, our framework computes feedback policies for concurrent multi-robot motion, focusing specifically on the execution uncertainty of autonomous skills in a shared workspace rather than general environmental uncertainty.

\begin{table}[t]
    \vspace{2pt}
    \caption{Comparison of multi-robot motion planning frameworks for skill execution and uncertainty.}
    \label{tab:related}
    \centering\scriptsize
    \setlength{\tabcolsep}{3.5pt}
    \begin{tabular}{lccll}
        \toprule
        & Multi- & Skill-Aware & \multirow{2}{*}{Uncertainty} & \multirow{2}{*}{Solution} \\
        & robot & planning & & \\
        \midrule
        \begin{tabular}{@{}l@{}}
            SMR~\cite{Alterovitz2008SMR}, iMDP~\cite{Huynh2012IncrementalSampling}, \\
            iPolicy~\cite{Zhao2024iPolicy}
        \end{tabular} 
        & & & own dynamics & policy \\
        ATM-MDP~\cite{Shah2020AnytimeIntegrated}            
        & & & task failure & policy \\
        TASP~\cite{Abbatematteo2024CIP}               
        & & \checkmark & --- & path \\
        MOSAIC~\cite{Mishani2026MOSAIC}             
        & & \checkmark & skill success & path \\
        APEX-MR~\cite{Huang2025APEXMR}            
        & \checkmark & \checkmark & timing delays & path\,+\,TPG \\
        MRMG~\cite{Hartmann2026SamplingBased}               
        & \checkmark & & --- & path \\
        \textbf{Ours} & \checkmark & \checkmark & spatial \& temporal & policy \\
        \bottomrule
    \end{tabular}
\end{table}

\section{Background: Multi-Robot Multi-Goal Planning}
\label{sec:background}
In many robotics applications, multiple robots share a workspace to complete a sequence of tasks as efficiently as possible. 
The multi-modal, multi-robot, multi-goal (MRMG) framework on which we build~\cite{Hartmann2026SamplingBased} treats these scenarios as a single centralized planning problem, avoiding the suboptimality of sequential decomposition at the cost of planning in a much higher dimensional space. 
We summarize only what our extension reuses and refer to the original work for further details.

\subsubsection{Tasks, modes, and the mode graph}
\label{sec:background:mrmg:tasks_modes}
A task consists of the assigned robots and the goal constraints that mark it complete.
Because completing tasks changes the physical attachments and collision geometry of the scene, the framework adopts a \emph{multi-modal} approach. 
A mode is defined as the tuple $m = (s, q_o)$, pairing the active task assignments $s \in \mathcal{S}$ with the relative object poses $q_o \in \mathcal{Q}_o$ (e.g., an object resting on a table versus attached to a gripper). 
This tuple defines a continuous subspace where the planning constraints are fixed and only the robot joints $q \in \mathcal{Q}_R$ are actuated. 
The valid progressions between modes are governed by an oracle $\mathcal{O}(m) \subseteq \mathcal{S}$ of admissible successor assignments. 
For dependency-constrained tasks, the oracle returns multiple admissible successors, allowing the planner to order independent tasks.

\subsubsection{Mode transitions}
\label{sec:background:mrmg:transitions}
To transition into a successor mode, the robots must reach a transition configuration where the current task's goal constraints are satisfied. 
Crossing this transition instantaneously completes the task, updating the assignment $s$ and potentially the attachments $q_o$. 
Transitions are therefore the place at which the discrete and continuous parts of the problem meet, and crucially, define the entry points from which skills are executed inside the respective skill modes (\cref{sec:concept}).

\subsubsection{Cost}
\label{sec:background:mrmg:cost}
Solutions are compared using a regularized maximum-makespan cost family:
\begin{equation}
    c(q_1, q_2) = (1-w) \max_{r} \lVert q_1^{r} - q_2^{r} \rVert_2
    + w \sum_{r} \lVert q_1^{r} - q_2^{r} \rVert_2 ,
    \label{eq:cost}
\end{equation}
summed over consecutive configurations of a path. 
The $\max$ term penalizes the largest per-robot displacement (the bottleneck of the makespan), while the weight $w \ll 1$ regularizes the large nullspace by penalizing unnecessary detours of the other robots. 

\subsubsection{Composite Planners}
\label{sec:background:mrmg:planners}
Within each mode, the MRMG framework plans across the joint configuration space of all robots $\mathcal{Q}_R = \prod_r \mathcal{Q}_r$.
Across modes, algorithms sample $(q,m)$ jointly, biasing mode sampling towards the not-yet-expanded frontier of the mode graph. 
Once an initial solution is found, the planners restrict subsequent sampling to locally informed subsets, and apply geometric shortcutting to incrementally refine path cost. 
We build upon the core planner architectures of this framework, namely prioritized planning (PP)~\cite{VanDenBerg2005Prioritized}, Probabilistic Roadmaps (\mbox{PRM*})~\cite{Kavraki1996PRM} and Rapidly-exploring Random Trees (\mbox{RRT*})~\cite{karaman2010incrementalsamplingbasedalgorithmsoptimal}.

\section{Skill-Aware MRMG Formulation}
\label{sec:concept}
To incorporate physical manipulation into the multi-agent planning problem, we replace the instantaneous geometric mode transitions of the standard MRMG framework (\cref{sec:background}) with skill modes, containing continuous physical skill executions. 
Upon entering a skill mode, the skill-executing robot acts autonomously.
The planner's role then shifts to coordinating the motion of the controllable robots to avoid collisions with the skill-executing robot throughout its execution.

\subsection{Skill Model \& Assumptions}
\label{sec:concept:skill_model}
To formalize this integration, we partition the set of all robots $R$ based on planner authority (i.e., whether a robot is steered directly by our planner or follows an autonomous skill controller).
In skill modes, the robot set is partitioned into autonomous skill-executing robots $R^{\text{skill}}$ and planner-steered robots $R^{\text{ctrl}}$.
In non-skill modes, all robots are controllable, forming the set $R^{\text{ctrl}} = R$.
Correspondingly, the composite configuration $q \in \mathcal{Q}_R$ factorizes as:
\begin{equation}
    q = (q^{\text{skill}}, q^{\text{ctrl}}) \in \mathcal{Q}_{R^{\text{skill}}} \times \mathcal{Q}_{R^{\text{ctrl}}} .
    \label{eq:factorization}
\end{equation}

Note that the \emph{skill/ctrl} labels describe a robot's role within the current mode, not a fixed identity: a robot executing a skill in one mode may act as a controllable robot in the next.

We model the skill as an autonomous, closed-loop controller with no external inputs.  
We assume access to a discrete-time transition function $f$ that simulates the skill's low-level execution, analogous to an environment transition model in reinforcement learning:
\begin{equation*}
    q^{\text{skill}}_{k+1} = f\left(q^{\text{skill}}_k, \text{env}_{\text{static}}, \omega_k\right) ,
    \label{eq:skill_step}
\end{equation*}
where the discrete step $k$ advances by the controller's step duration $\Delta t$, $\text{env}_{\text{static}}$ provides static scene context (e.g., target object frames and static obstacle geometry for the controller's kinematics), and $\omega_k$ represents an arbitrary stochastic disturbance. 
This formulation supports arbitrary, multimodal skill distributions and natively captures the stochastic behavior of the controller.

The underlying controller $f$ guarantees self-collision avoidance and collision-free motion against static obstacles, and we assume its execution is fully observable. 
It does \emph{not} account for the controllable robots, so inter-robot collision avoidance remains the responsibility of the centralized planner.
Note that $f$ can represent multiple robots executing individual skills concurrently, or even a collaborative multi-robot skill involving several manipulators in $R^{\text{skill}}$, without altering our formulation.
 
In the special case where the stochastic disturbance is negligible ($\omega_k=0$), the outcome distribution collapses to a single nominal trajectory with a zero-volume uncertainty envelope. 
This naturally reduces the problem to deterministic planning.

\section{Deterministic Skill Planning}
\label{sec:deterministic}
Before addressing stochastic skills, we establish a foundation by integrating deterministic skills into the MRMG framework. 
This provides a baseline for scenarios where uncertainty is negligible. 
We adapt three planner classes to evaluate how skill integration performs across fundamentally different search strategies.

\subsubsection{Prioritized Planning (PP)}
\label{sec:deterministic:pp}
PP assigns a priority to each robot and plans their paths sequentially. 
During a skill mode, the deterministic skill trajectory is unrolled over time, and lower-priority controllable robots treat the skill-executing robot as a known, time-varying dynamic obstacle.
As discussed in \cref{sec:experiments:deterministic_baseline}, this approach is inherently suboptimal and often incomplete.

\subsubsection{Tree-Based Planning (\mbox{RRT*})}
\label{sec:deterministic:rrt}
Tree-based planners search forward in time from an initial state, making skill integration a natural extension.
Instead of unrolling full skill chains upfront, the tree grows by incrementally querying the skill controller $f$ to advance $q^{\text{skill}}$ while concurrently steering $q^{\text{ctrl}}$ toward sampled targets. 

\subsubsection{Probabilistic Roadmaps (\mbox{PRM*})}
\label{sec:deterministic:prm}
Standard \mbox{PRM*} connects sampled configurations using undirected edges.
To enforce the same forward progression as in \mbox{RRT*}, \mbox{PRM*} embeds each skill as an unrolled sequence of discrete states that must be traversed from start to finish.
For each state along this sequence, candidate configurations for $R^{\text{ctrl}}$ are sampled and combined with the corresponding state of $R^{\text{skill}}$ to form a layer of composite vertices.
Edges connect vertices from consecutive layers, so $R^{\text{ctrl}}$ can move as $R^{\text{skill}}$ progresses.
This creates a directed, layered network where a graph search (A*) finds the optimal collision-free path through this network for $R^{\text{ctrl}}$ while respecting the skill rollout.

Across the asymptotically optimal planners, standard enhancements such as informed sampling and path shortcutting are retained. 
The shortcutting algorithm is specifically constrained to interpolate only non-skill branches, preserving the physical validity of the skill executions.

\section{Stochastic Skill Planning}
\label{sec:stochastic}
In real-world execution, manipulation skills are rarely perfectly deterministic. 
As the skill-executing robot $R^{\text{skill}}$ operates stochastically, its execution yields an uncertain trajectory $\tau^{\text{skill}} = (q^{\text{skill}}_0, \dots, q^{\text{skill}}_K)$ of potentially variable horizon $K$ (e.g., due to retry loops during grasping). 
Given this uncertainty, finding a single open-loop path is insufficient. 
Instead, the planner must compute a closed-loop policy $\pi^\star$ that prescribes how the controllable robots $R^{\text{ctrl}}$ should dynamically react to avoid collisions, and minimize expected cost:
\begin{equation}
    \pi^\star = \argmin_{\pi} \; \mathbb{E} \left[ \sum_{k=0}^{T-1} c_{\text{time}}(q_k, q_{k+1}) \right] ,
    \label{eq:ssp}
\end{equation}
where $T$ is the total execution horizon.
While deterministic planners evaluate stage cost $c$ geometrically (see \cref{eq:cost}), manipulation skills have variable execution times that cannot be captured by geometric distance alone.
To accurately penalize robots idling for delayed skills, the stage cost $c_{\text{time}}$ evaluates the elapsed time between consecutive states:
\begin{equation}
    c_{\text{time}}(q_k, q_{k+1}) = 
    \begin{cases}
        \Delta t & \text{in skill modes,} \\
        c(q_k, q_{k+1}) / v_{\max} & \text{in non-skill modes,}
    \end{cases}
    \label{eq:cost_time}
\end{equation}
where $\Delta t$ is the policy decision period during skill execution, while $v_{\max}$ converts controllable-robot motion into time. 
We solve the resulting MDP over the multi-modal roadmaps and compare its reactive policy with the conservative open-loop baseline.

\subsection{Conservative Baseline}
\label{sec:stochastic:conservative}
As a baseline, we construct an open-loop multi-robot path $\sigma = (q_0, \dots, q_T)$ spanning all modes. 
Outside of skill modes, all robots are controllable, and the planner operates in the full composite space.
During a skill mode, where $R^{\text{skill}}$ is dictated by its stochastic controller, the planner steers the remaining controllable robots $R^{\text{ctrl}}$ while blindly advancing $R^{\text{skill}}$ along its expected nominal trajectory.

Because an open-loop plan commits to a rigid sequence of configurations, it must be planned conservatively.
To robustly approximate safety across the true distribution of skill outcomes, we sample $N_{\text{MC}}$ Monte Carlo rollouts of the skill. 
A single static envelope around the skill trajectory that needs to be avoided at all times would be overly conservative in our shared-workspace environments, since it could block regions that $R^{\text{ctrl}}$ must enter even when the skill robot is not there at the corresponding time.
We therefore check the controllable robot against the skill realizations at their corresponding time steps.
To ensure temporal and spatial safety against these realizations, we extend the baseline trajectory to the maximum observed horizon $K_{\max}$ and collision-check each waypoint $q^{\text{ctrl}}_k$ against the cloud of $N_{\text{MC}}$ realizations at step $k$.
\begin{equation}
    \big(q^{\text{skill}, (i)}_k,\, q^{\text{ctrl}}_k\big) \in \mathcal{Q}_{\text{free}}
    \quad \forall\, i = 1, \dots, N_{\text{MC}} .
    \label{eq:tubecheck}
\end{equation}
This allows $R^{\text{ctrl}}$ to safely maneuver within the shared workspace when timing permits.

\subsection{Reactive Planner}
\label{sec:stochastic:reactive}
The conservative baseline and reactive planner differ fundamentally in the feedback available during execution:
\begin{align*}
    \text{Conservative Plan:} \quad & q_{k+1}^{\text{ctrl}} = \sigma^{\text{ctrl}}(k+1) , 
    \\
    \text{Reactive Policy:} \quad & q_{k+1}^{\text{ctrl}} = \pi\big(m,\, q_k^{\text{ctrl}},\, q_k^{\text{skill}}\big) . 
\end{align*}

Here, $\sigma^{\mathrm{ctrl}}$ denotes the controllable-robot component of the composite open-loop path.
The skill-executing component is fixed by the nominal skill rollout.

Notice that the conservative open-loop plan depends solely on the time index $k$, forcing $R^{\text{ctrl}}$ into a worst-case schedule. 
In contrast, the reactive policy conditions its actions directly on the real-time observed state of both $R^{\text{ctrl}}$ and $R^{\text{skill}}$, dynamically adapting the motions of $R^{\text{ctrl}}$ to the realized progression.

To compute the reactive policy $\pi^\star$, we formulate and solve a discrete MDP through the following systematic stages.

\subsubsection{Roadmaps}
\label{sec:stochastic:reactive:roadmaps}
In non-skill modes, the existing roadmap spans the full composite configuration space.
In each skill mode, we instead build a lower-dimensional roadmap over $\mathcal{Q}_{R^{\mathrm{ctrl}}}$, while sampled skill rollouts represent the stochastic state of the skill-executing robots.
Edge lengths are limited to $\rho=v_{\max}\Delta t$ so that one action fits within a single policy period.

\subsubsection{Skill Outcome Binning}
\label{sec:stochastic:reactive:binning}
To solve the discrete MDP without full workspace discretization, we compress $N_{\text{MC}}$ skill rollouts into a finite transition model.
At each epoch $t_k = k\Delta t$, continuous configurations $q^{\text{skill}}$ are quantized into spatio-temporal nature states $z$:
\begin{equation}
    z = \Big(k,\, \big\lfloor q^{\text{skill}}(t_k)/\delta \big\rceil\Big),
    \ \ 
    P(z' \mid z) = \frac{n(z \!\to\! z')}{\sum_{z''} n(z \!\to\! z'')} ,
    \label{eq:nature}
\end{equation}
where $\delta$\ is the discretization resolution and $P(z' \mid z)$ are empirical transition probabilities.
Spatial-only binning aliases fast and slow skill realizations at shared locations, breaking the Markov property. 
Appending the decision epoch $k$ restores the Markov property and forms a directed acyclic graph (DAG), eliminating cycles and allowing exact backward Bellman backups.
Bins $z$ also store medoids $q_z$ and representative rollout observations for explicit multi-sample collision checking against $R^{\text{ctrl}}$, alongside terminal exit configurations $e_z$ for mode transitions.

\subsubsection{Markov Decision Process}
\label{sec:stochastic:reactive:mdp}
We define the MDP state $x$ as the tuple $(m, u, z)$, where $m$ denotes the current mode, $u \in \mathcal{V}_{\text{ctrl}}$ is a vertex in the $R^{\text{ctrl}}$ roadmap, and $z \in \mathcal{Z}$ is the discrete nature state of $R^{\text{skill}}$. 
For each skill mode, this joint state space can be viewed as a product of the controllable roadmap and the skill's nature-transition graph, analogous to the tensor-product roadmap explored by discrete RRT~\cite{solovey2016drrt}.
Each action $a \in \mathcal{A}(u) = \mathcal{N}(u) \cup \{u\}$ directs the controllable robot to transition to a neighbor vertex $v \in \mathcal{N}(u)$ in the roadmap, or to hold its current configuration ($v=u$).
The stage cost $g(u,z,a,z')$ corresponds to the time-based stage cost $c_{\text{time}}$ defined in \cref{eq:cost_time}.
To enforce collision avoidance, we pre-compute a static Boolean mask, marking as infeasible any state where the composite configuration $(q_z, q_u)$ results in a collision between $R^{\text{ctrl}}$ and $R^{\text{skill}}$.

\subsubsection{Value Iteration}
\label{sec:stochastic:reactive:value_iteration}
The policy is computed via a backward pass through the skill's directed graph, propagating expected cost-to-go values from terminal states back to the start.
For skill modes, the expected cost-to-go $V(u,z)$ for $R^{\text{ctrl}}$ is computed using the Bellman equation:
\begin{equation}
    V(u,z) = \min_{a \in \mathcal{A}(u)} \sum_{z'} P(z' \mid z)
    \big[\, g(u,z,a,z') + W(a,z,z') \,\big] .
    \label{eq:bellman}
\end{equation}
The continuation value $W(a,z,z')$ branches based on whether the skill continues executing or terminates.
Because termination yields no future active state $z'$, the exit value depends only on the state $z$ from which the skill terminated:
\begin{equation}
    W(a,z,z') =
    \begin{cases}
        V(a, z') & \text{if skill continues} , \\
        V^{\text{exit}}(a, z) & \text{if skill terminates} .
    \end{cases}
    \label{eq:continuation}
\end{equation}
When a skill terminates, the exact timing and final configuration are uncertain. 
The exit value $V^{\text{exit}}(a,z)$ dynamically evaluates the cost of seamlessly transitioning the final realized composite configuration into the successor mode to continue the multi-agent plan:
\begin{equation}
    V^{\text{exit}}(a,z) = \min_{n} \Big[\, c\big((e_z, q_a),\, q_n\big) + V(n) \,\Big] .
    \label{eq:exitvalue}
\end{equation}
This exit connection frees $R^{\text{ctrl}}$ from having to perfectly align with a pre-sampled transition node when the skill unpredictably ends.

For non-skill modes, when the environment transitions deterministically ($P=1$), the stochastic branching collapses into a deterministic shortest-path search solved via Dijkstra's algorithm.

\subsubsection{Policy Shortcutting \& Anytime Batched Refinement}
\label{sec:stochastic:reactive:policy_shortcutting}
Rather than constructing a full multi-modal roadmap prior to value iteration, which risks exhausting the planning budget before finding a feasible policy, we adopt an anytime batched approach. 
We first grow the sampled multimodal roadmap until a feasible path from the start state to a terminal mode is found.
Once an initial path to the goal is found, we extract an initial policy and enter an iterative refinement loop, alternating between incremental roadmap expansion, value iteration, and policy shortcutting.
By simulating the closed-loop policy against sampled skill realizations, we generate likely rollouts and apply straight-line shortcuts exclusively to the $R^{\text{ctrl}}$ segments, as $R^{\text{skill}}$ is strictly governed by the autonomous controller.
Validated shortcuts are then injected directly back into the roadmap.
This densifies the roadmap precisely where the policy requires it, enabling subsequent Bellman backups to lower the expected cost $V(\text{start})$ across successive planning batches until the budget expires.

\subsubsection{Execution}
\label{sec:stochastic:reactive:execution}
During execution, continuous observations $(q^{\text{skill}}, q^{\text{ctrl}})$ are quantized into the corresponding MDP state $(z, u)$, and the action prescribed by $\pi^\star$ is retrieved in $\mathcal{O}(1)$ time.
If execution noise results in an observation outside the Monte Carlo support, it is projected onto the nearest visited state within the current epoch layer $k$.

\begin{figure*}[t]
    \centering
    \begin{minipage}[t]{0.24\textwidth}
        \centering
        \includegraphics[width=\linewidth]{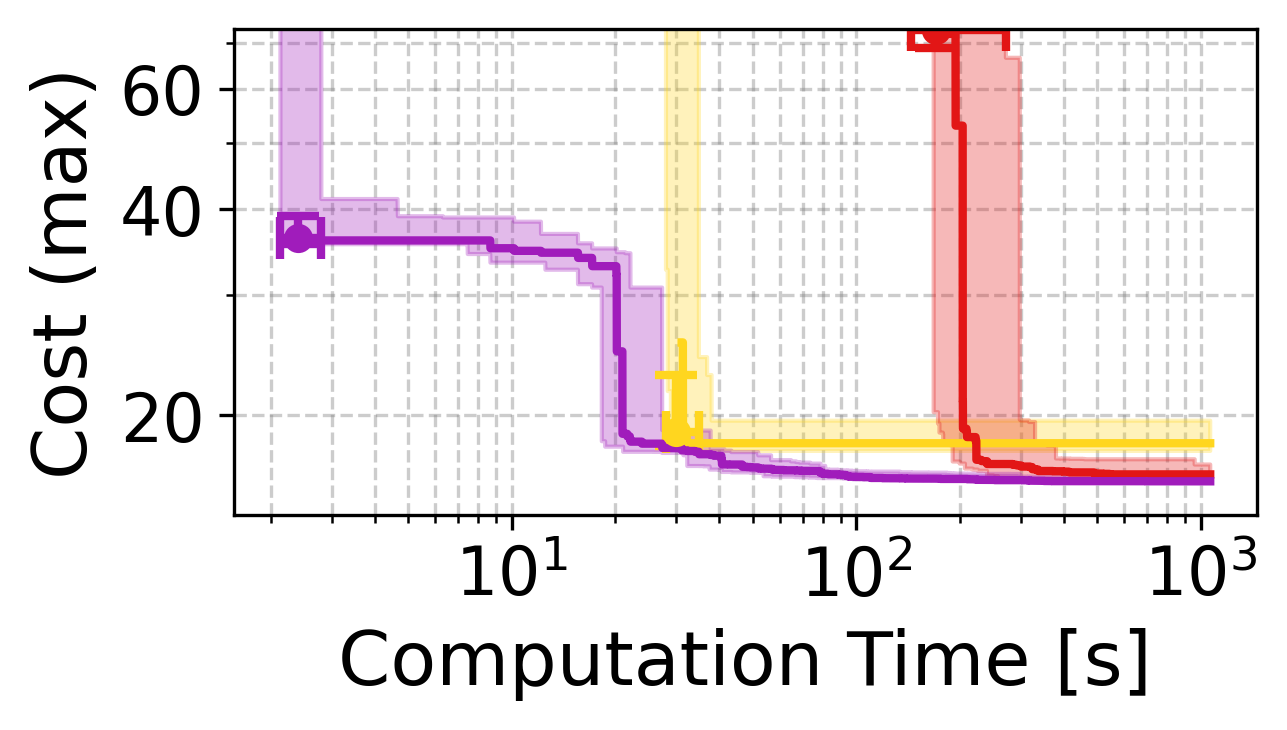}\\
        \vspace{2pt}
        \includegraphics[width=0.60\linewidth]{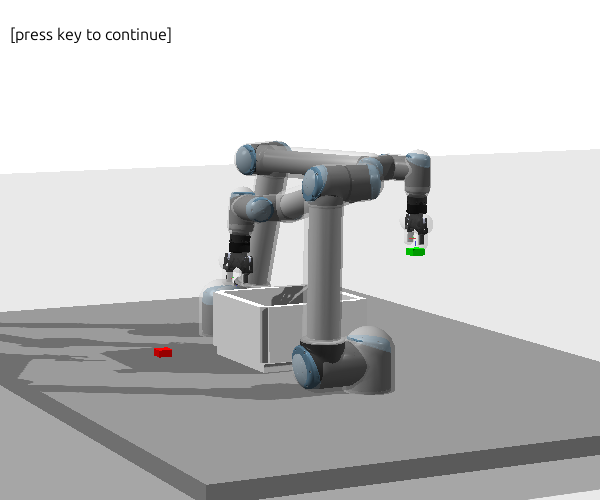}\\
        \vspace{3pt}
        {\footnotesize\textit{(a) Bin picking.}}
    \end{minipage}\hfill
    \begin{minipage}[t]{0.24\textwidth}
        \centering
        \includegraphics[width=\linewidth]{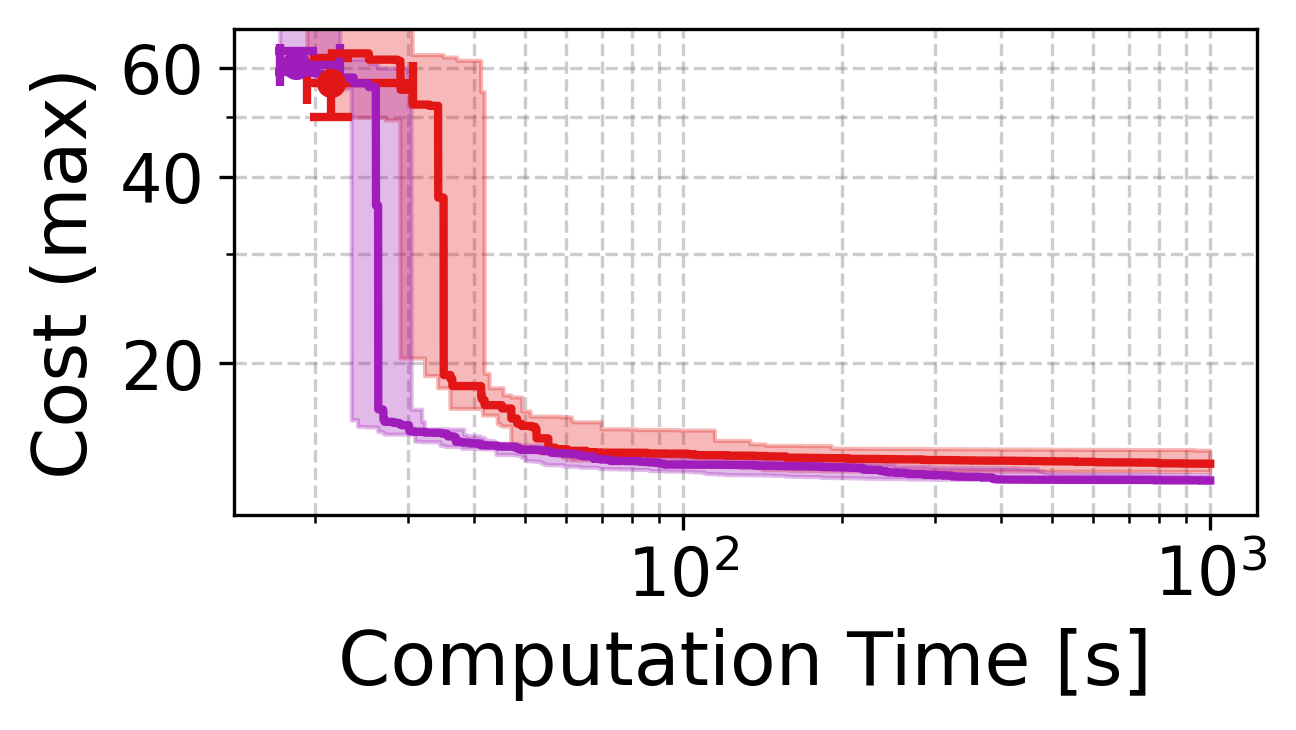}\\
        \vspace{2pt}
        \includegraphics[width=0.60\linewidth]{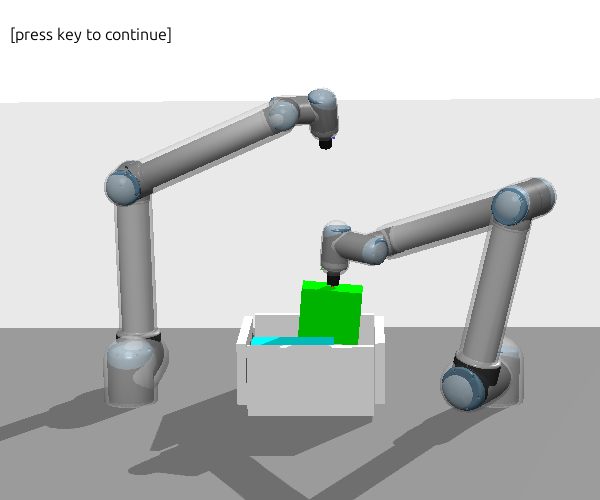}\\
        \vspace{3pt}
        {\footnotesize\textit{(b) Bin packing.}}
    \end{minipage}\hfill
    \begin{minipage}[t]{0.24\textwidth}
        \centering
        \includegraphics[width=\linewidth]{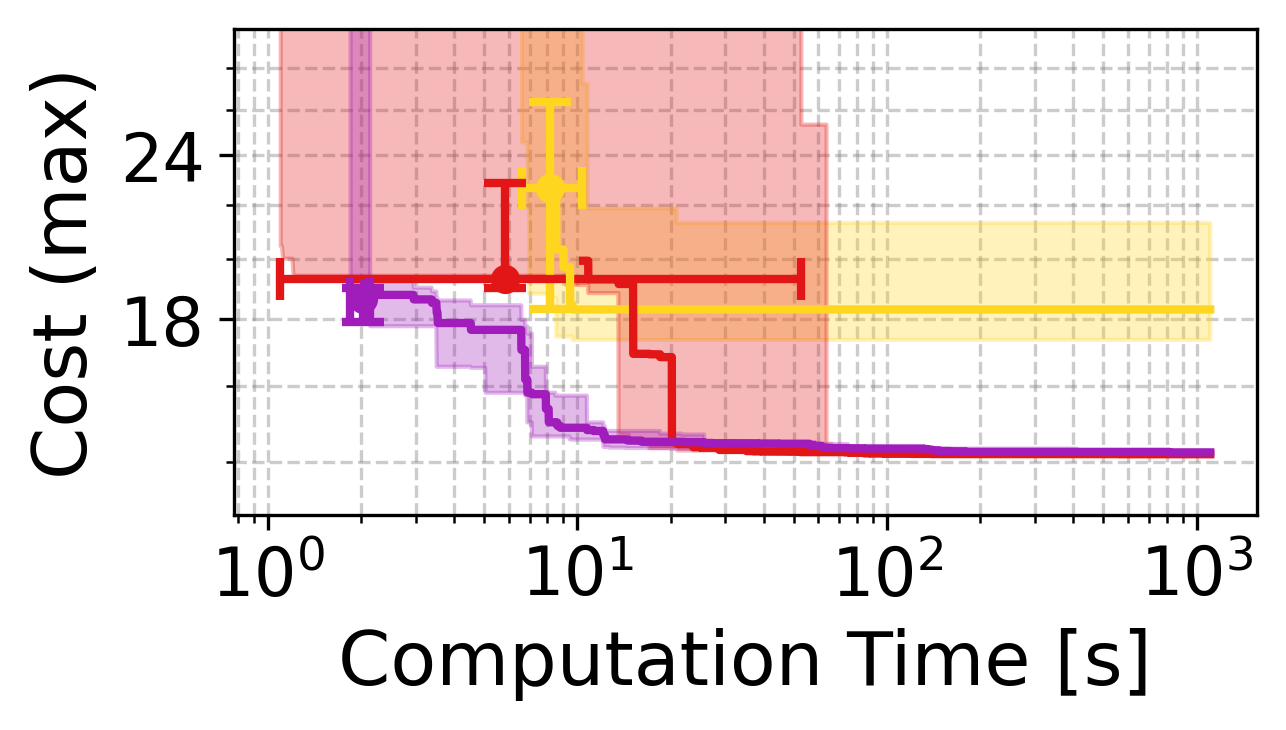}\\
        \vspace{2pt}
        \includegraphics[width=0.60\linewidth]{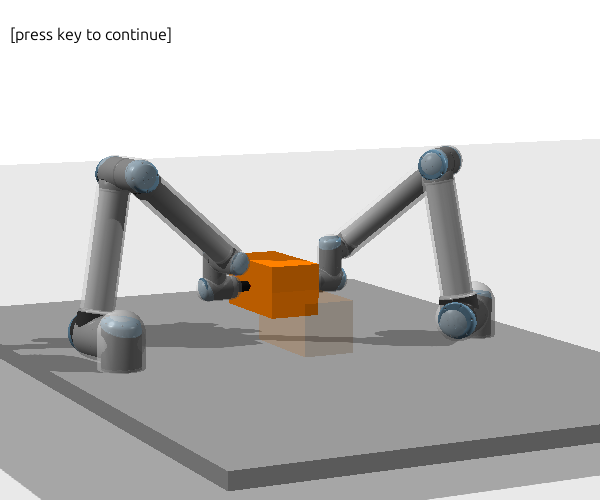}\\
        \vspace{3pt}
        {\footnotesize\textit{(c) Dual-arm transport.}}
    \end{minipage}\hfill
    \begin{minipage}[t]{0.24\textwidth}
        \centering
        \includegraphics[width=\linewidth]{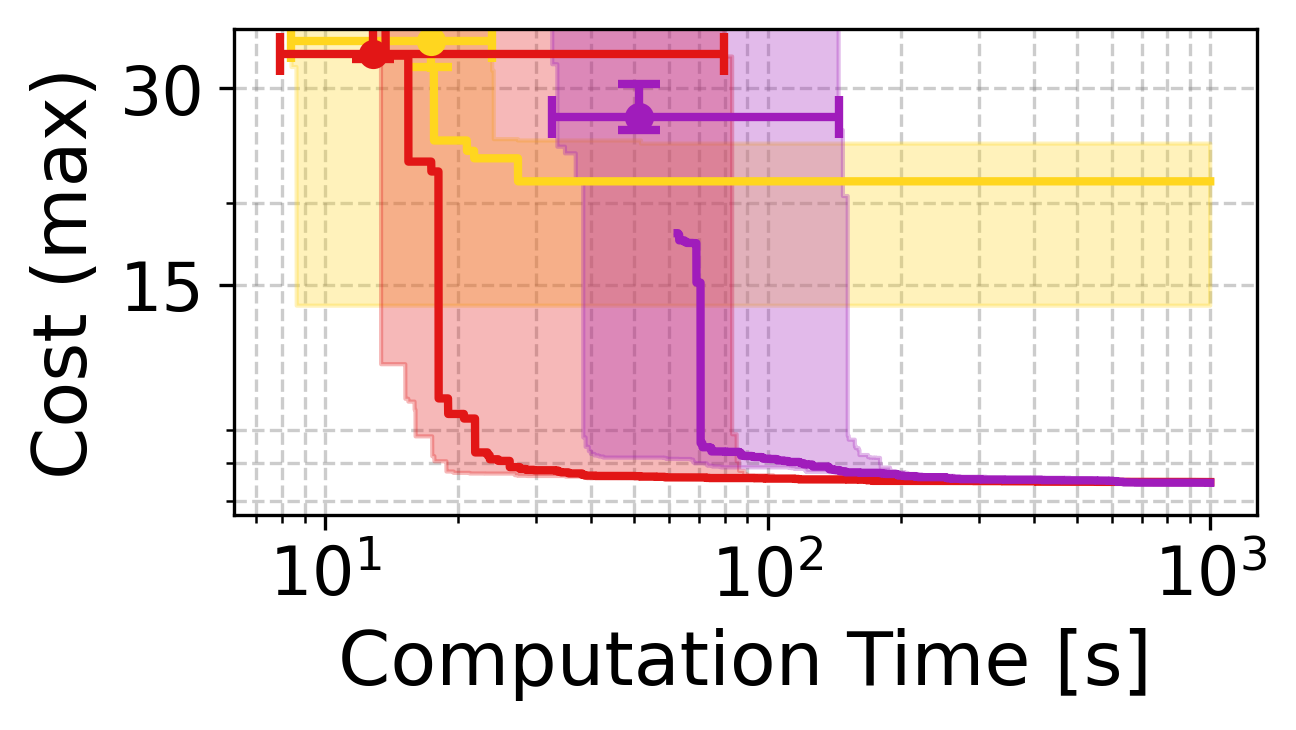}\\
        \vspace{2pt}
        \includegraphics[width=0.60\linewidth]{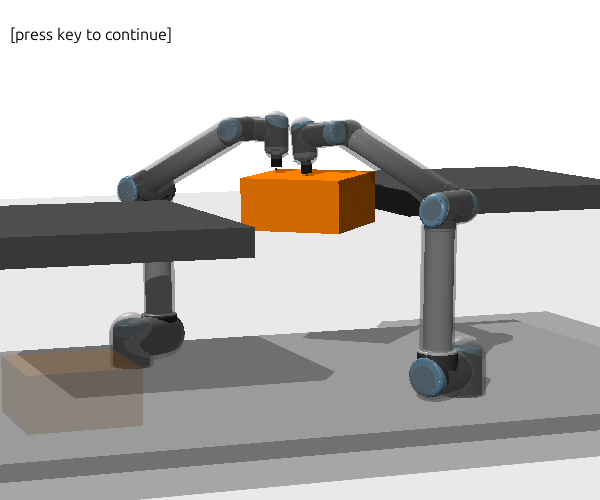}\\
        \vspace{3pt}
        {\footnotesize\textit{(d) Handover.}}
    \end{minipage}

    \vspace{6pt}
    \centerline{\begin{tikzpicture}
\begin{axis} [
  width=\linewidth,
  height=0.5\textwidth,
  unbounded coords=jump,
  xtick align=inside,
  ytick align=inside,
  anchor=north,
  hide axis,
  clip=false,
  xmajorgrids,
  ymajorgrids,
  major grid style={densely dotted, black!20},
  xmin=0,
  xmax=10,
  ymin=0,
  ymax=10,
  xlabel style={font=\footnotesize},
  xticklabel style={font=\footnotesize},
  ylabel style={font=\footnotesize},
  yticklabel style={font=\footnotesize},
  legend style={anchor=south, legend cell align=left, legend columns=3, at={(axis cs:5, 6)}, font=\small},
  legend image code/.code={\draw[#1] (0cm,0cm) -- (0.3cm,0cm);}
]

\addlegendimage{colorPP, line width=1.0pt}
\addlegendentry{PP}

\addlegendimage{colorPRM, line width=1.0pt}
\addlegendentry{PRM*}

\addlegendimage{colorRRT, line width=1.0pt}
\addlegendentry{RRT*}

\end{axis}
\end{tikzpicture}}
    \caption{Deterministic skill-planning results for four multi-agent manipulation environments.
    Each panel combines the cost convergence plot with an illustration of the corresponding environment. 
    The curves show the best solution cost found as computation proceeds, with planner colors and line styles defined by the shared legend.}
    \label{fig:deterministic_results}
\end{figure*}

\begin{figure*}[t]
    \centering
    \begin{minipage}[t]{0.49\textwidth}
        \vspace{-10pt}
        \centering
        \subfloat[\label{fig:mega_2d_analytics:sq_cons_down}]{\begin{minipage}[t][0.25\linewidth][b]{0.25\linewidth}\centering\includegraphics[width=\linewidth]{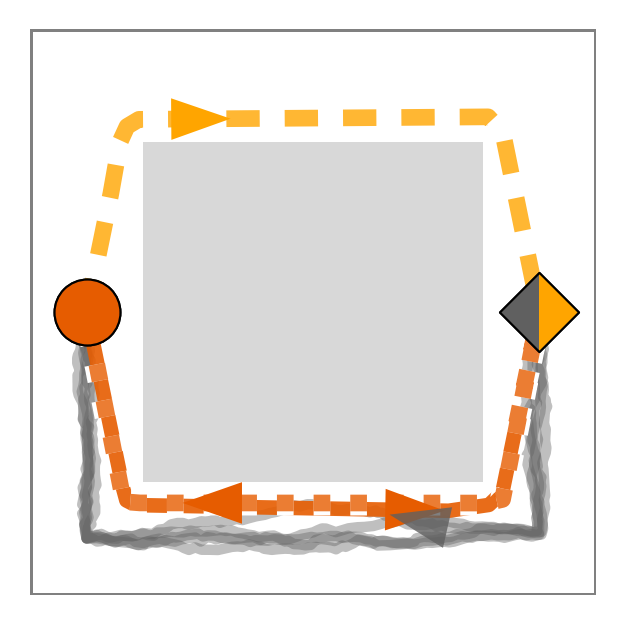}\end{minipage}}\hfill
        \subfloat[\label{fig:mega_2d_analytics:sq_cons_up}]{\begin{minipage}[t][0.25\linewidth][b]{0.25\linewidth}\centering\includegraphics[width=\linewidth]{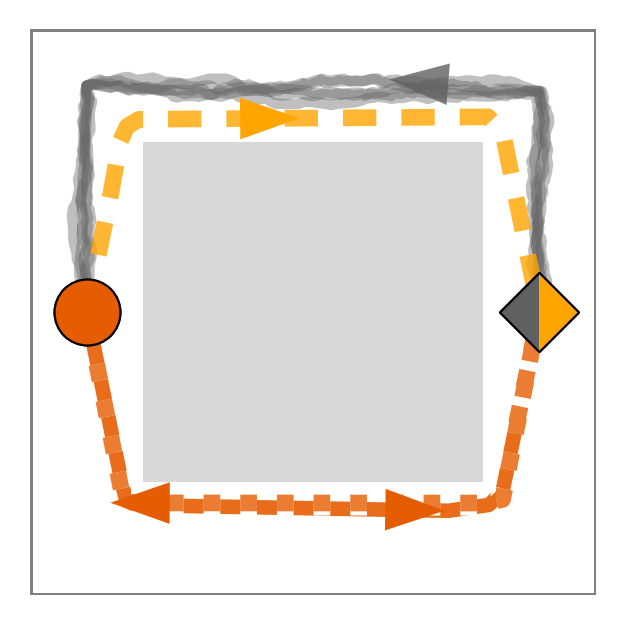}\end{minipage}}\hfill
        \subfloat[\label{fig:mega_2d_analytics:sq_react_down}]{\begin{minipage}[t][0.25\linewidth][b]{0.25\linewidth}\centering\includegraphics[width=\linewidth]{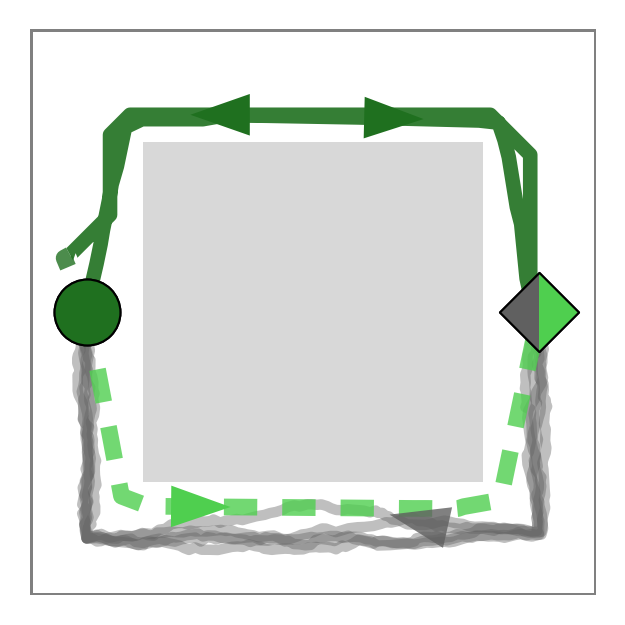}\end{minipage}}\hfill
        \subfloat[\label{fig:mega_2d_analytics:sq_react_up}]{\begin{minipage}[t][0.25\linewidth][b]{0.25\linewidth}\centering\includegraphics[width=\linewidth]{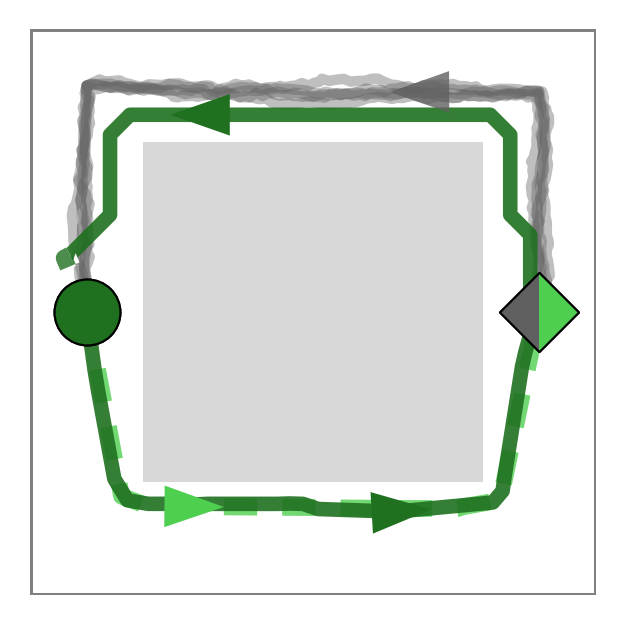}\end{minipage}}
        \vspace{0pt}
        \subfloat[\label{fig:mega_2d_analytics:rect_cons_down}]{\begin{minipage}[t][0.25\linewidth][b]{0.25\linewidth}\centering\includegraphics[width=\linewidth]{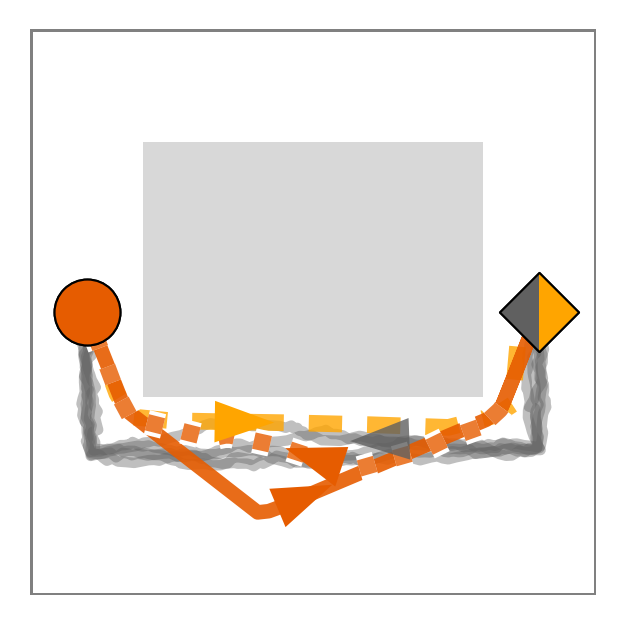}\end{minipage}}\hfill
        \subfloat[\label{fig:mega_2d_analytics:rect_cons_up}]{\begin{minipage}[t][0.25\linewidth][b]{0.25\linewidth}\centering\includegraphics[width=\linewidth]{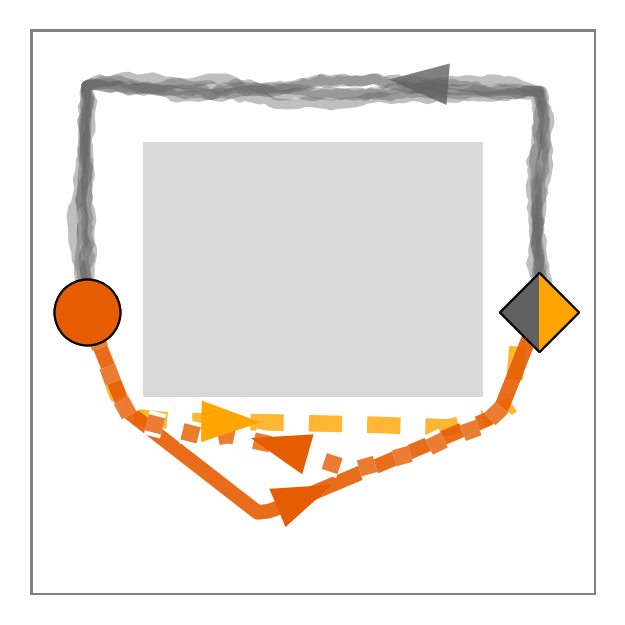}\end{minipage}}\hfill
        \subfloat[\label{fig:mega_2d_analytics:rect_react_down}]{\begin{minipage}[t][0.25\linewidth][b]{0.25\linewidth}\centering\includegraphics[width=\linewidth]{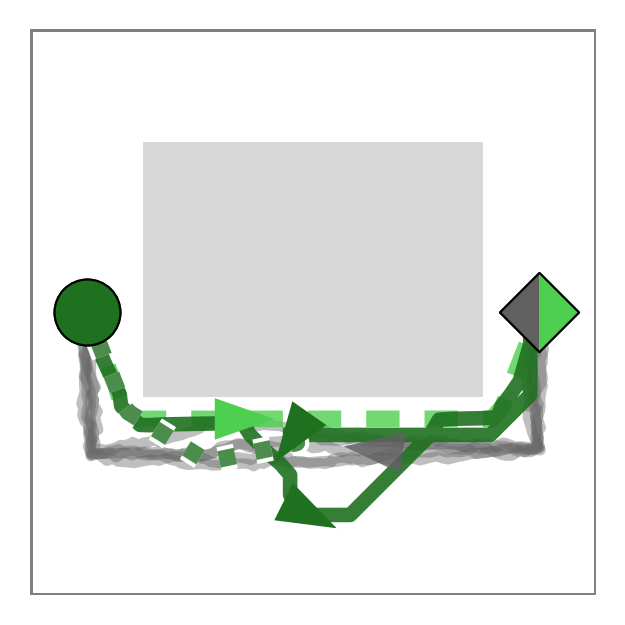}\end{minipage}}\hfill
        \subfloat[\label{fig:mega_2d_analytics:rect_react_up}]{\begin{minipage}[t][0.25\linewidth][b]{0.25\linewidth}\centering\includegraphics[width=\linewidth]{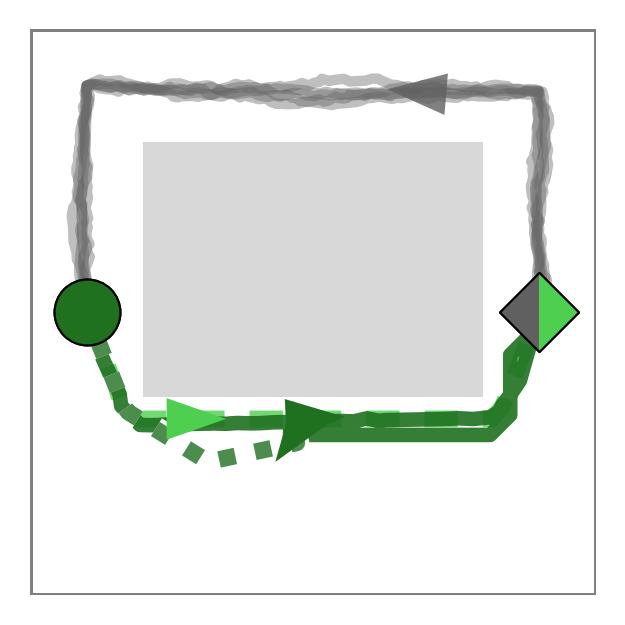}\end{minipage}}
        \vspace{14pt}
    \end{minipage}
    \hfill
    \begin{minipage}[t]{0.49\textwidth}
        \vspace{-10pt}
        \centering
        \subfloat[\label{fig:mega_2d_analytics:sq_cost}]{\begin{minipage}[t][0.25\linewidth][b]{0.5\linewidth}\centering\includegraphics[width=\linewidth]{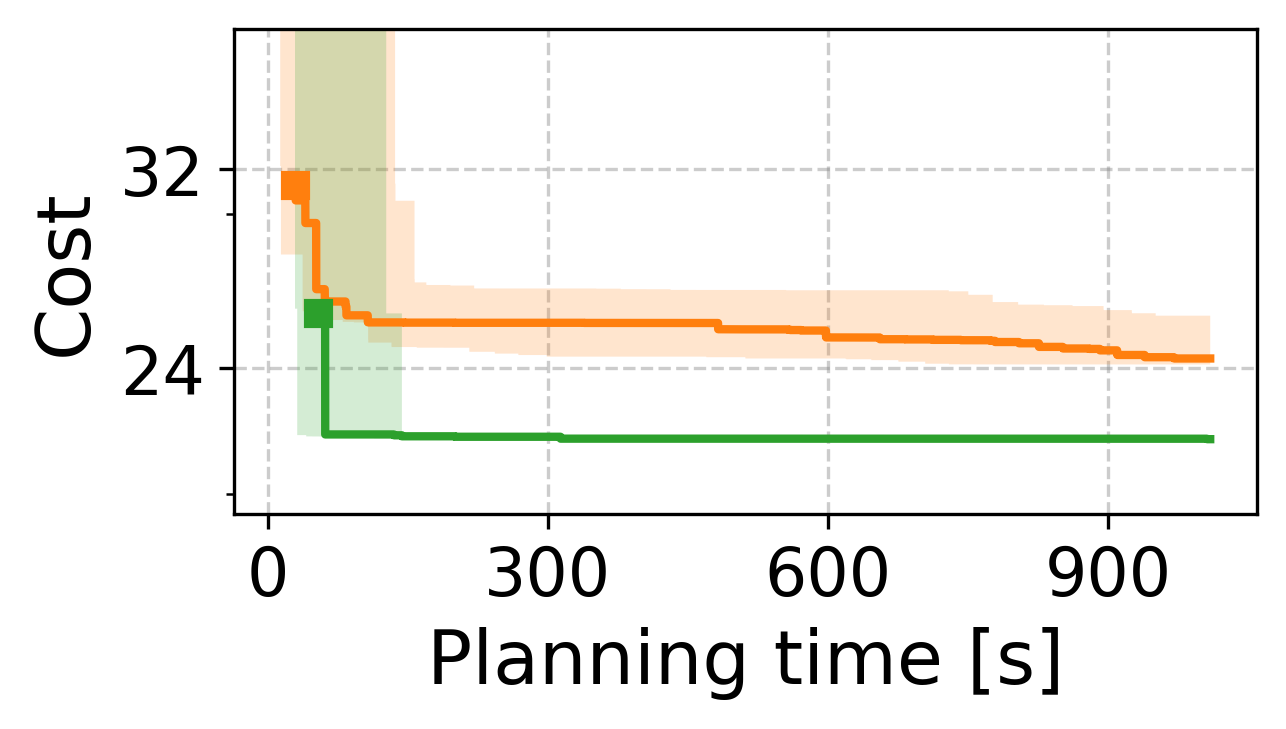}\end{minipage}}\hfill
        \subfloat[\label{fig:mega_2d_analytics:sq_epdf}]{\begin{minipage}[t][0.25\linewidth][b]{0.5\linewidth}\centering\includegraphics[width=\linewidth]{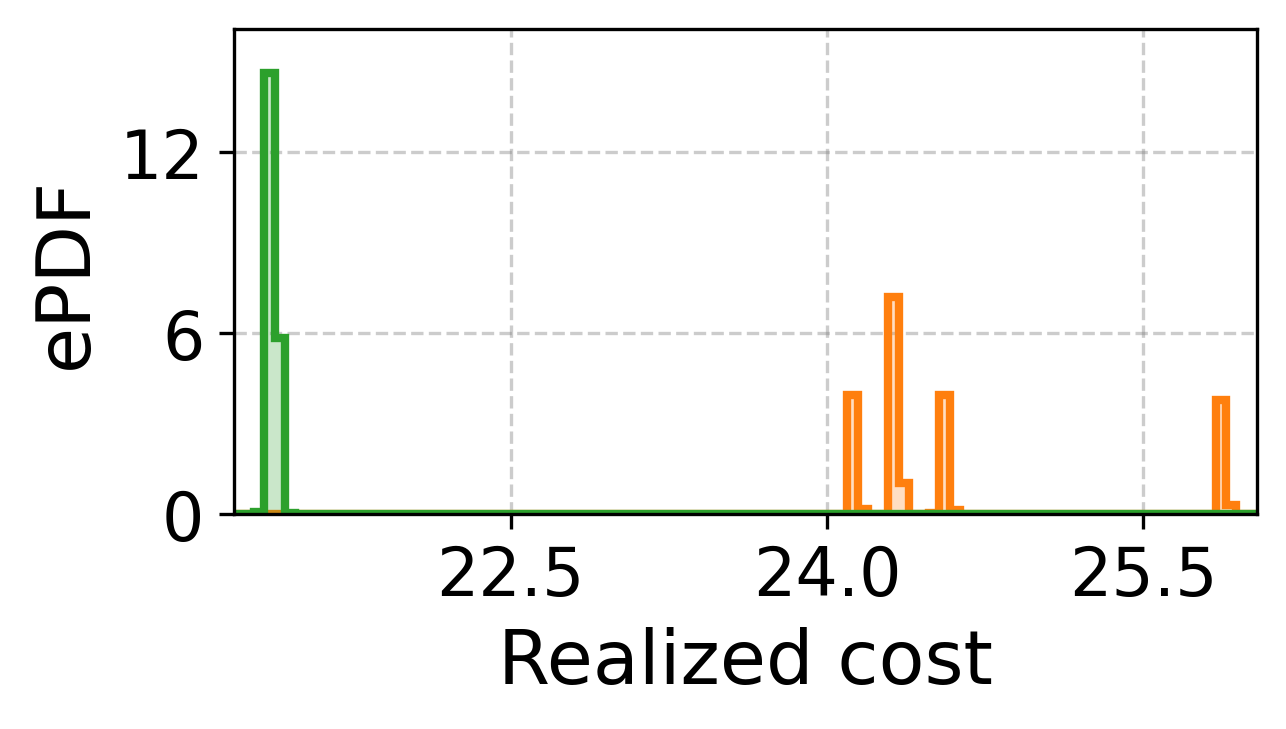}\end{minipage}}
        \vspace{0pt}
        \subfloat[\label{fig:mega_2d_analytics:rect_cost}]{\begin{minipage}[t][0.25\linewidth][b]{0.49\linewidth}\centering\includegraphics[width=\linewidth]{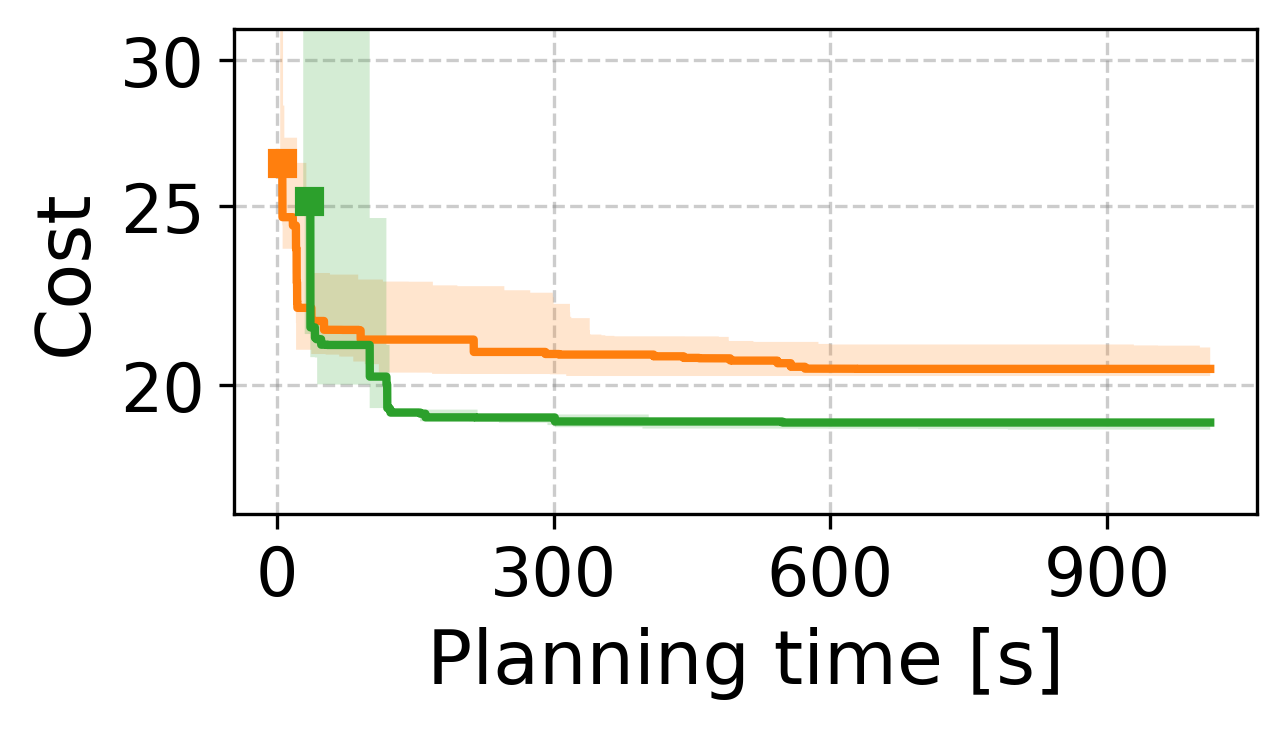}\end{minipage}}\hfill
        \subfloat[\label{fig:mega_2d_analytics:rect_epdf}]{\begin{minipage}[t][0.25\linewidth][b]{0.507\linewidth}\centering\includegraphics[width=\linewidth]{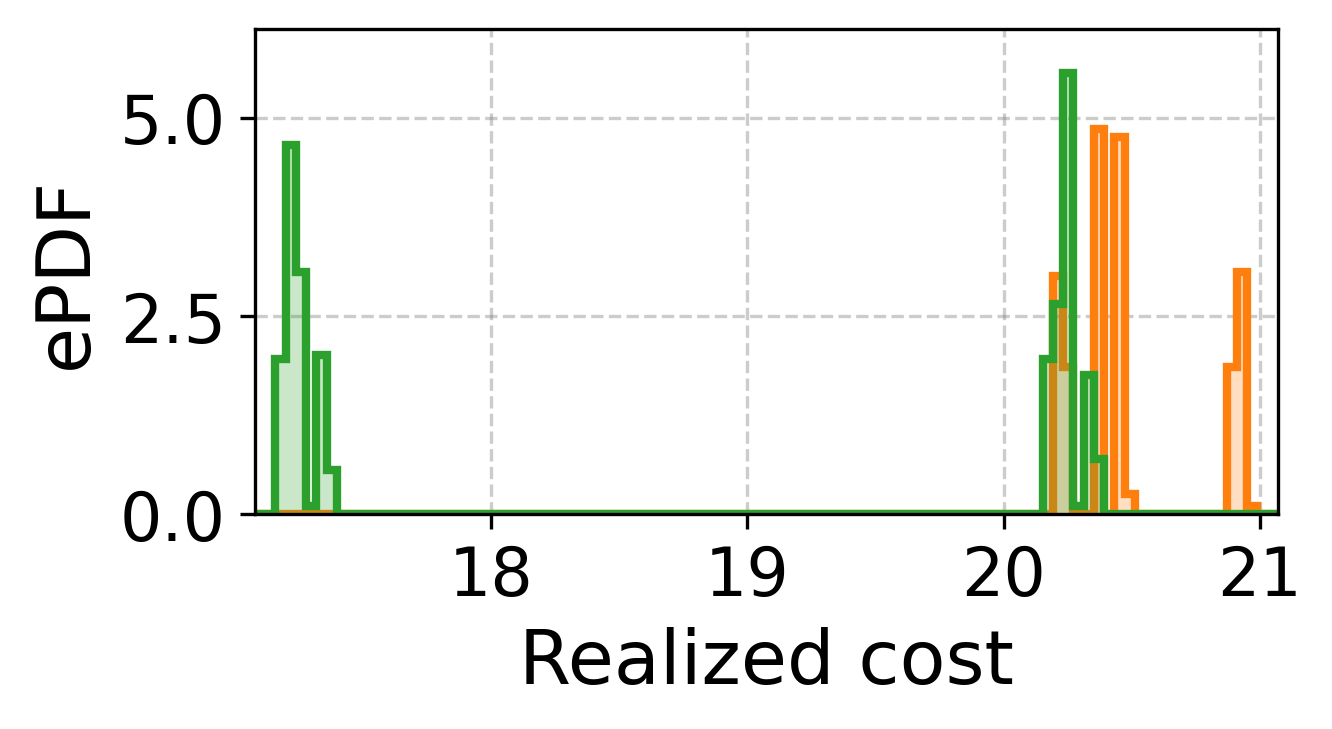}\end{minipage}}
        \vspace{5pt}
    \end{minipage}

    \vspace{-6pt}
    \centerline{  {\small
  \setlength{\fboxsep}{5pt}
  \fbox{%
  \begin{tabular}{@{}r@{\ }c@{\ }c@{\ }c@{\ }c@{\ }c@{\ }c@{}}

  \textcolor{colorConservative}{Conservative:}
  &
  \begin{tikzpicture}[baseline=-0.55ex]
    \draw[colorStartGoal, line width=1.6pt] (0,0) -- (0.32cm,0);
  \end{tikzpicture}
  $R^{\mathrm{skill}}$
  &
  $\rightarrow$
  &
  \begin{tikzpicture}[baseline=-0.55ex]
    \draw[colorConservativeSkill, densely dotted, line width=1.6pt]
      (0,0) -- (0.32cm,0);
  \end{tikzpicture}
  $R^{\mathrm{ctrl}}$
  &
  \begin{tikzpicture}[baseline=-0.55ex]
    \draw[colorConservativeCtrl, line width=1.6pt] (0,0) -- (0.32cm,0);
  \end{tikzpicture}
  $R^{\mathrm{ctrl}}$
  &
  $\rightarrow$
  &
  \begin{tikzpicture}[baseline=-0.55ex]
    \draw[colorConservativeCtrl, densely dotted, line width=1.6pt]
      (0,0) -- (0.32cm,0);
  \end{tikzpicture}
  $R^{\mathrm{ctrl}}$
  \\[-1pt]

  \textcolor{colorReactive}{Reactive:}
  &
  \begin{tikzpicture}[baseline=-0.55ex]
    \draw[colorStartGoal, line width=1.6pt] (0,0) -- (0.32cm,0);
  \end{tikzpicture}
  $R^{\mathrm{skill}}$
  &
  $\rightarrow$
  &
  \begin{tikzpicture}[baseline=-0.55ex]
    \draw[colorReactiveSkill, densely dotted, line width=1.6pt]
      (0,0) -- (0.32cm,0);
  \end{tikzpicture}
  $R^{\mathrm{ctrl}}$
  &
  \begin{tikzpicture}[baseline=-0.55ex]
    \draw[colorReactiveCtrl, line width=1.6pt] (0,0) -- (0.32cm,0);
  \end{tikzpicture}
  $R^{\mathrm{ctrl}}$
  &
  $\rightarrow$
  &
  \begin{tikzpicture}[baseline=-0.55ex]
    \draw[colorReactiveCtrl, densely dotted, line width=1.6pt]
      (0,0) -- (0.32cm,0);
  \end{tikzpicture}
  $R^{\mathrm{ctrl}}$

  \end{tabular}%
  }}}

    \caption{Planar coordination and execution statistics for (\textbf{top}) the square-island and (\textbf{bottom}) rectangular-island environments.
    The left panels show representative executions, where circles and rhombi mark starts and terminal goals and arrows indicate motion direction.
    During skill execution, $R^{\text{skill}}$ is gray and $R^{\text{ctrl}}$ is solid in the planner color. After skill completion, both trajectories are dotted in that color.
    The right panels show cost convergence and ePDFs of realized cost for the conservative and reactive planners.}
    \label{fig:mega_2d_analytics}
\end{figure*}

\section{Experiments and Results}
\label{sec:experiments}

\subsection{Experimental Setup}
\label{sec:experiments:setup}
We evaluate our framework in three parts. 
First, we evaluate deterministic planning in four multi-robot manipulation environments with two 6-DoF arms, resulting in a 12-DoF composite robot configuration space $\mathcal{Q}_R \subseteq \mathbb{R}^{12}$. 
Second, we introduce stochastic skills in planar environments with two 2-DoF robots ($\mathcal{Q}_R \subseteq \mathbb{R}^{4}$) to compare conservative and reactive planning. 
Finally, we evaluate scalability on an eight-object stacking task with two to four 6-DoF manipulators, resulting in composite configuration spaces with 12--24 DoF.

Unless stated otherwise, each planning configuration is evaluated over 5 independent planning runs with a maximum planning time of 1000\,s per run.
For stochastic experiments, each resulting path or policy is additionally evaluated over 100 independent skill realizations.
Variation across planning runs reflects differences in the sampled planning process and resulting solutions, whereas variation across executions reflects stochastic skill outcomes.
In convergence plots, solid curves show the pointwise median across planning runs and shaded regions show the corresponding pointwise interquartile range.

All experiments are implemented in Python and executed on an Intel Core i7 CPU with 16\,GB of RAM. %

\subsection{Deterministic Planning (Baseline)}
\label{sec:experiments:deterministic_baseline}
We first compare PP, \mbox{PRM*}, and \mbox{RRT*} to assess how different planner architectures handle deterministic skill execution.

As shown in \cref{fig:deterministic_results}, \mbox{RRT*} typically finds initial solutions fastest and converges to optimal costs.
\mbox{PRM*} achieves comparable solutions, but constructing a multi-modal roadmap across the entire composite space can be less efficient for this single-query setting than the goal-directed search of \mbox{RRT*}. 
In contrast, the suboptimal PP is more sensitive to the strict priority ordering and frequently fails or produces substantially higher costs in tightly coupled workspaces.

These results establish the deterministic planning baseline and motivate the stochastic experiments, where such open-loop plans fail under variable skill execution because they cannot adapt to the realized skill outcome.

\subsection{Conservative vs. Reactive Planning}
\label{sec:experiments:stochastic_lowdim}
We next illustrate the behavioral differences between conservative and reactive planning in three planar environments with stochastic skills.
The examples highlight how online adaptation affects coordination and realized cost. 
We compare the resulting coordination strategies through their motions, cost evolution, and realized-cost distributions.
Numerical improvements refer to the mean realized makespan over successful executions.

\subsubsection{Square Environment (Symmetric Corridors)}
\label{sec:experiments:stochastic_lowdim:square}
Both robots navigate to the opposite side and return. 
During the initial skill mode, the rhombus robot executes a stochastic outbound skill as $R^{\text{skill}}$, while the disk robot acts as $R^{\text{ctrl}}$ and maneuvers around it. 
After the skill terminates, both robots become controllable and execute their return motions.

As shown in \cref{fig:mega_2d_analytics:sq_cons_down,fig:mega_2d_analytics:sq_cons_up}, the conservative planner slows $R^{\text{ctrl}}$ to let the robots pass safely, which is only possible in the corners because of the narrow corridors. 
This delay prevents $R^{\text{ctrl}}$ from reaching its checkpoint before the skill terminates. 
In contrast, the reactive policy uses the unoccupied corridor without waiting, allowing $R^{\text{ctrl}}$ to reach its checkpoint and begin its return motion while the skill is still executing.
This reduces the mean realized makespan by 13.3\% over successful executions (\cref{fig:mega_2d_analytics:sq_cost}).
The reactive costs are tightly concentrated, whereas the conservative costs form several narrow clusters caused by differences between independently generated open-loop plans (\cref{fig:mega_2d_analytics:sq_epdf}).

\subsubsection{Rectangle Environment (Asymmetric Corridors)}
\label{sec:experiments:stochastic_lowdim:rectangle}
The rectangle environment uses the same task but asymmetric corridors, producing distinct spatial routes and stochastic skill durations.

Both planners identify and attempt to exploit the shorter corridor, but their execution differs. 
The conservative baseline plans for the worst-case timing and spatial outcome, whereas the reactive policy adapts its motion to the realized skill trajectory.
The reactive executions are bimodal, with a low-cost mode around 17.3\,s and a high-cost mode around 20.2\,s.
Although the two modes occur with similar frequency, the reactive planner reduces the mean realized makespan by 8.1\% over successful executions (\cref{fig:mega_2d_analytics:rect_cost,fig:mega_2d_analytics:rect_epdf}).

\begin{figure}[!t]
    \centering
    \includegraphics[width=\columnwidth]{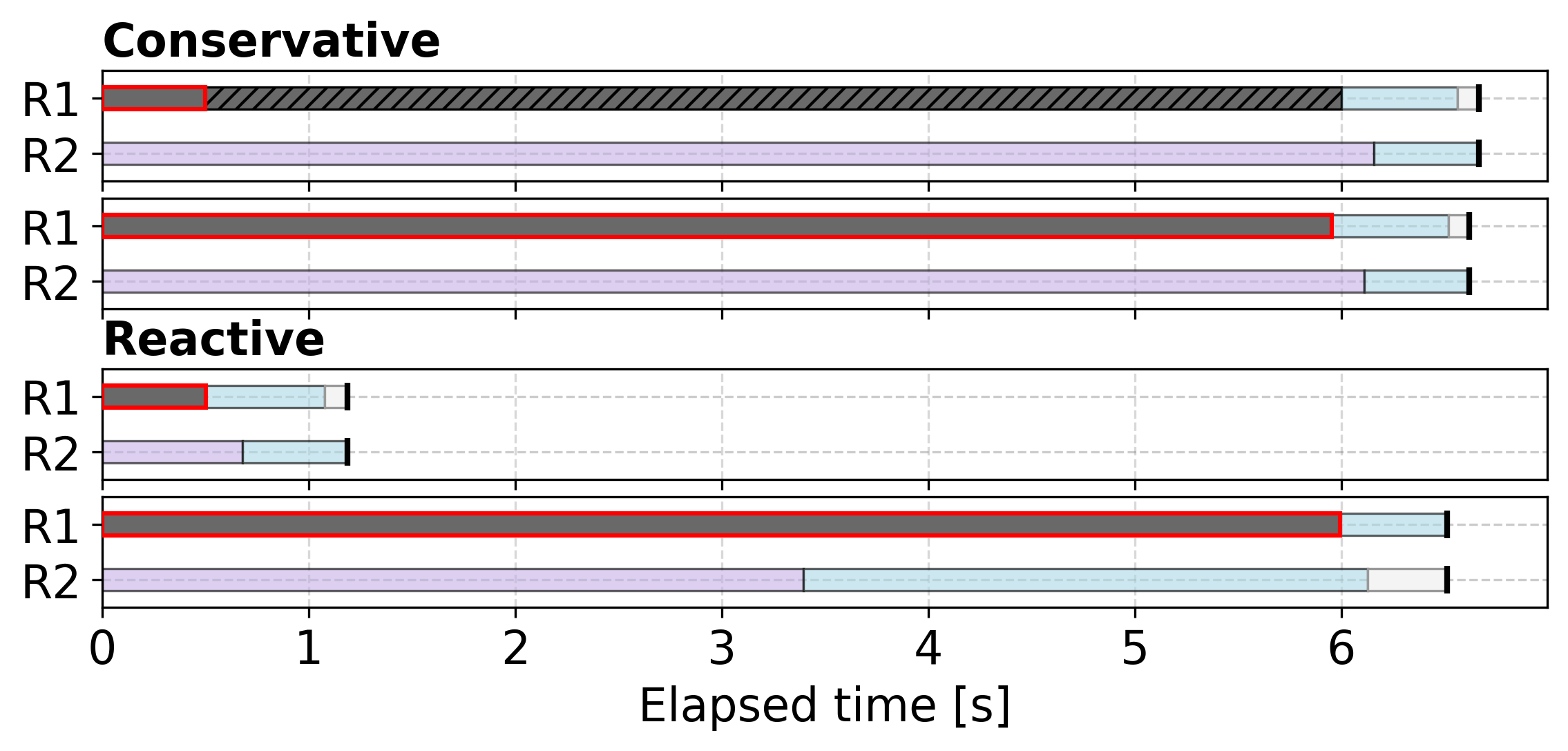}
    \centerline{\begin{tikzpicture}
\begin{axis} [
  width=\linewidth,
  height=0.5\textwidth,
  unbounded coords=jump,
  xtick align=inside,
  ytick align=inside,
  anchor=north,
  hide axis,
  clip=false,
  xmajorgrids,
  ymajorgrids,
  major grid style={densely dotted, black!20},
  xmin=0,
  xmax=10,
  ymin=0,
  ymax=10,
  xlabel style={font=\footnotesize},
  xticklabel style={font=\footnotesize},
  ylabel style={font=\footnotesize},
  yticklabel style={font=\footnotesize},
  legend style={anchor=south, legend cell align=left, legend columns=5, at={(axis cs:5, 6)}, font=\small}
]

\addlegendimage{legend image code/.code={
  \fill[colorSkillFill] (0cm,-0.08cm) rectangle (0.16cm,0.08cm);
  \draw[red, line width=0.9pt] (0cm,-0.08cm) rectangle (0.16cm,0.08cm);
}}
\addlegendentry{Skill}

\addlegendimage{legend image code/.code={
  \fill[colorSkillFill] (0cm,-0.08cm) rectangle (0.16cm,0.08cm);
  \draw[colorBoxBorder, line width=0.5pt, pattern=north east lines, pattern color=black] (0cm,-0.08cm) rectangle (0.16cm,0.08cm);
}}
\addlegendentry{Wait}

\addlegendimage{legend image code/.code={
  \fill[colorPhasePurple] (0cm,-0.08cm) rectangle (0.16cm,0.08cm);
  \draw[colorBoxBorder, line width=0.5pt] (0cm,-0.08cm) rectangle (0.16cm,0.08cm);
}}
\addlegendentry{Visit}

\addlegendimage{legend image code/.code={
  \fill[colorPhaseBlue] (0cm,-0.08cm) rectangle (0.16cm,0.08cm);
  \draw[colorBoxBorder, line width=0.5pt] (0cm,-0.08cm) rectangle (0.16cm,0.08cm);
}}
\addlegendentry{Return}

\addlegendimage{legend image code/.code={
  \fill[colorPhaseGray] (0cm,-0.08cm) rectangle (0.16cm,0.08cm);
  \draw[colorTerminalBorder, line width=0.5pt] (0cm,-0.08cm) rectangle (0.16cm,0.08cm);
}}
\addlegendentry{Terminal}

\end{axis}
\end{tikzpicture}}
    \caption{Representative execution schedules for the shared-point task.
    The upper pair shows the conservative planner and the lower pair the reactive planner. 
    Within each pair, the upper and lower schedules correspond to the short and long realizations, respectively.
    Horizontal bars denote each robot's task intervals, hatched bars indicate waiting.}
    \label{fig:2d_shared_point_gantt}
\end{figure}

\subsubsection{Shared Point Environment}
\label{sec:experiments:stochastic_lowdim:shared_point}
Both robots must navigate to a shared central point and return, while $R^{\text{skill}}$ may execute its skill quickly or slowly.
As illustrated in the Gantt chart (\cref{fig:2d_shared_point_gantt}), the conservative baseline commits to a worst-case schedule $K_{\max}$, forcing $R^{\text{skill}}$ to wait even when its skill finishes early.
The reactive policy instead adapts online and uses the task dependencies during slow executions to trigger a mid-skill mode switch, allowing $R^{\text{ctrl}}$ to reach its goal before the skill terminates.
This adaptation reduces the mean realized makespan by 43.4\%.

\begin{table}[t]
    \vspace{2pt}
    \centering
    \caption{Planning and execution results for the eight-object stacking task.
    `Sequence' and `Dependency' denote the task-order settings, $R\in\{2,3,4\}$ is the number of robots, and $P\in\{\colorsq{colorConservative}\,\text{conservative},\colorsq{colorReactive}\,\text{reactive}\}$ denotes the planner.
    $\bar{t}_{\mathrm{first}}$ is the mean time to the first finite path or policy, and $C_{\mathrm{med}}$ is the median realized makespan over successful executions. 
    `Plan succ.' is the fraction of planning runs that produce a valid path or policy, while `Exec. succ.' is the fraction of attempted stochastic executions that reach the terminal task state.}
    \label{tab:stochastic_stacking}
    \vspace{2pt}
    \footnotesize
    \setlength{\tabcolsep}{2pt}
    \renewcommand{\arraystretch}{0.96}
    \resizebox{\columnwidth}{!}{%
    \begin{tabular}{c c c c c c c c}
        \toprule
        \textbf{Setting} & \textbf{P} & \textbf{R}
        & \begin{tabular}[b]{@{}c@{}}\textbf{Modes}\\$|\mathcal M|$\end{tabular}
        & \begin{tabular}[b]{@{}c@{}}$\boldsymbol{\bar{t}_{\mathrm{first}}}$\\{[s]}\end{tabular}
        & \begin{tabular}[b]{@{}c@{}}\textbf{Plan succ.}\\ {[\%]}\end{tabular}
        & \begin{tabular}[b]{@{}c@{}}\textbf{Exec. succ.}\\ {[\%]}\end{tabular}
        & \begin{tabular}[b]{@{}c@{}}$\boldsymbol{C_{\mathrm{med}}}$\\{[s]}\end{tabular}\\
        \midrule
        \multirow{6}{*}{\rotatebox{90}{\textbf{Sequence}}}
        & \colorsq{colorConservative} & \multirow{2}{*}{2} & \multirow{2}{*}{27} & 17.4 & 100 & 100 & 61.4 \\
        & \colorsq{colorReactive} & & & 80.6 & 100 & 100 & 46.6 \\
        \cmidrule(lr){2-8}
        & \colorsq{colorConservative} & \multirow{2}{*}{3} & \multirow{2}{*}{28} & 21.4 & 100 & 100 & 61.5 \\
        & \colorsq{colorReactive} & & & 99.8 & 100 & 100 & 48.8 \\
        \cmidrule(lr){2-8}
        & \colorsq{colorConservative} & \multirow{2}{*}{4} & \multirow{2}{*}{29} & 23.1 & 100 & 100 & 62.6 \\
        & \colorsq{colorReactive} & & & 133.6 & 100 & 100 & 50.7 \\
        \midrule
        \multirow{6}{*}{\rotatebox{90}{\textbf{Dependency}}}
        & \colorsq{colorConservative} & \multirow{2}{*}{2} & \multirow{2}{*}{73} & 23.9 & 100 & 80 & 47.4 \\
        & \colorsq{colorReactive} & & & 375.6 & 100 & 69 & 47.7 \\
        \cmidrule(lr){2-8}
        & \colorsq{colorConservative} & \multirow{2}{*}{3} & \multirow{2}{*}{200} & 19.9 & 100 & 100 & 46.8 \\
        & \colorsq{colorReactive} & & & 257.8 & 100 & 59 & 56.0 \\
        \cmidrule(lr){2-8}
        & \colorsq{colorConservative} & \multirow{2}{*}{4} & \multirow{2}{*}{535} & 19.6 & 100 & 100 & 52.9 \\
        & \colorsq{colorReactive} & & & 226.1 & 60 & 89 & 57.7 \\
        \bottomrule
    \end{tabular}%
    }

\end{table}

\begin{figure}[t]
    \centering
    \subfloat[\label{fig:stochastic_stacking_eval:seq_cost}]{\includegraphics[width=0.49\columnwidth]{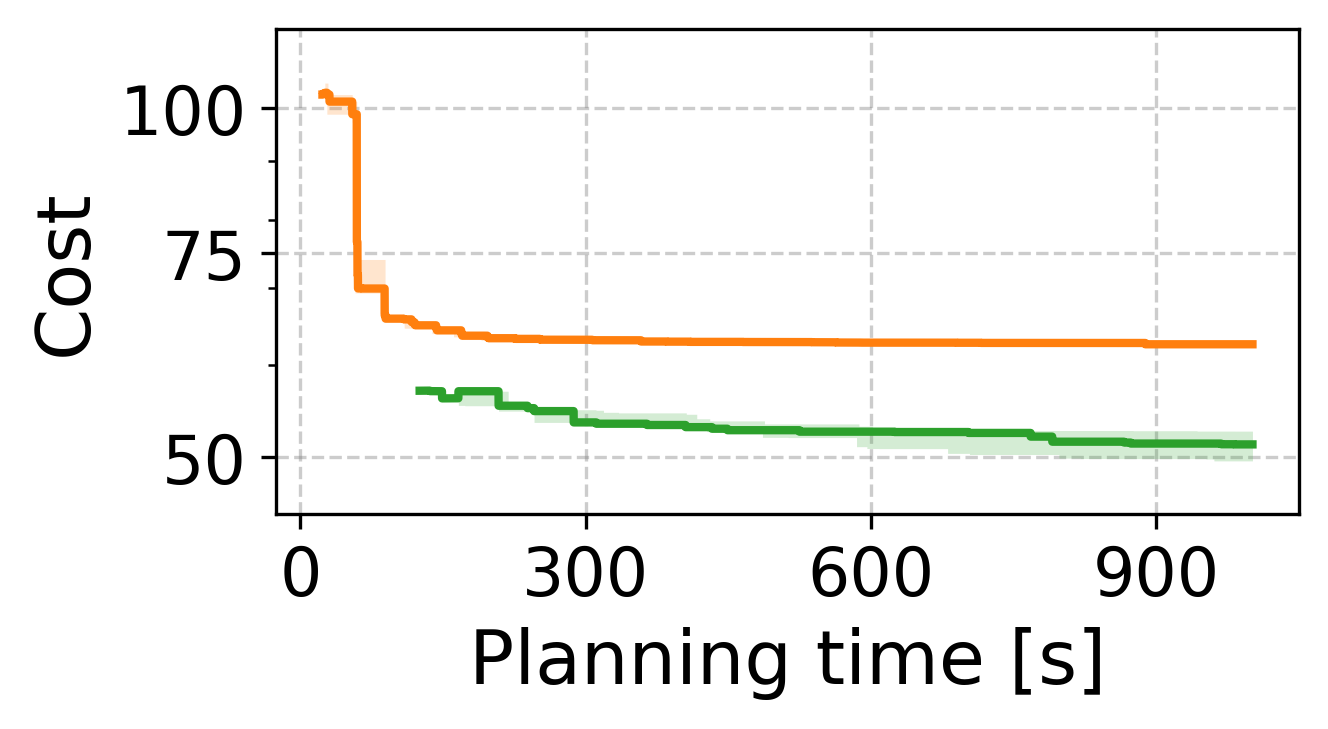}}
    \subfloat[\label{fig:stochastic_stacking_eval:seq_epdf}]{\includegraphics[width=0.49\columnwidth]{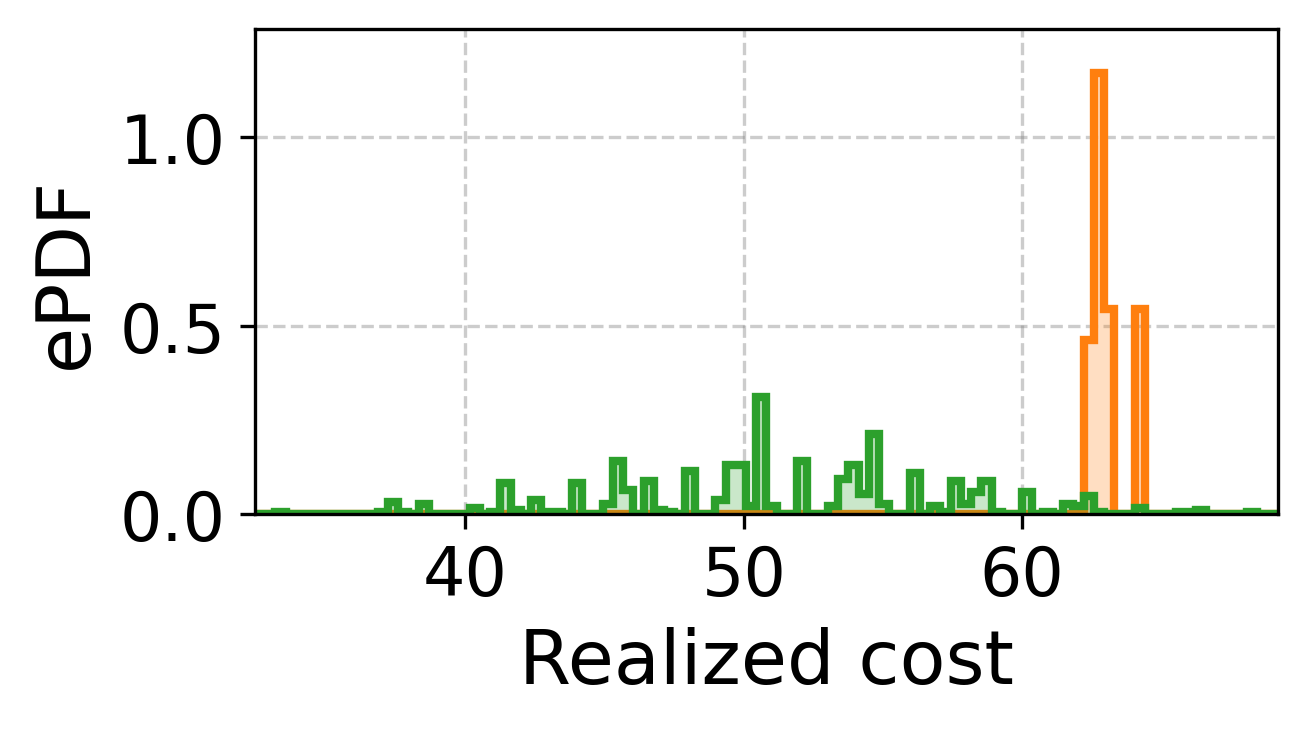}}\hfill
    \caption{Planning-cost convergence and realized-makespan distributions for the prescribed-sequence 4R--8-object instance for \colorsq{colorConservative}\,\text{conservative} and \colorsq{colorReactive}\,\text{reactive} planners.
    The left panel shows the planning-cost evolution for the conservative and reactive planners, while the right panel shows the ePDF of realized makespan over successful stochastic executions of the resulting solutions.}
    \label{fig:stochastic_stacking_eval}
\end{figure}

\subsection{High-Dimensional Robot-Arm Simulation}
\label{sec:experiments:stochastic_highdim}
Finally, we evaluate scalability under two task-order settings: a prescribed sequence and a dependency graph. 
In both settings, each pick skill lasts either 1 or 3\,s with equal probability, allowing us to assess how stochastic execution and task-order complexity affect planning and execution.

With the task order fixed, adding robots increases the continuous dimension without introducing additional task-order branches. 
The mode count therefore grows only slightly with the number of robots rather than combinatorially as it does for the dependency graph formulation.
The exact counts are reported in \cref{tab:stochastic_stacking}.

\subsubsection{Prescribed sequence}
Under the prescribed sequence, both planners achieve 100\% planning and execution success. 
The conservative planner finds an initial solution 4.6--5.8$\times$ faster, whereas the reactive planner reduces the median realized makespan by 19.0--24.1\% by adapting to the realized duration of the stochastic pick skill instead of reserving the worst case throughout. 
This trade-off is reflected in the convergence and realized-cost distributions shown in \cref{fig:stochastic_stacking_eval}.

\subsubsection{Dependency graph}
With task dependencies, the number of discovered modes during planning increases from 73 to 200 to 535 as robots grow from two to four.
Unlike the prescribed sequence, planning and execution success now depend on both the planner and robot count, as shown in \cref{tab:stochastic_stacking}. 
The results expose the increasing difficulty of covering and connecting the stochastic branches required for reactive policy construction.

This setting is more demanding for reactive planning because the policy must provide valid continuations across many stochastic branches, whereas the conservative planner only needs one feasible route through the task graph. 
The dependency results therefore expose, as expected, a scalability bottleneck in achieving sufficient coverage and connectivity in the roadmap--MDP, rather than in the online feedback mechanism itself.
This higher difficulty means that in some rollouts, no solution is found for the problem, and it is possible that a plan execution fails.

\section{Conclusion}
\label{sec:conclusion}
We presented a sampling-based framework for multi-modal, multi-robot, multi-goal motion planning that integrates stochastic manipulation skills at planning time.
The framework represents skill execution and geometric motion in a common composite space and supports both conservative open-loop planning and reactive closed-loop coordination.

The conservative planner handles uncertainty through worst-case scheduling, which can cause spatial detours and temporal idling.
To overcome these limitations, we presented an anytime batched reactive planner that formulates and incrementally solves an MDP over the multi-modal roadmap.
The resulting policy adapts the controllable robots' motions to the observed skill execution using lightweight online action lookup.

The planar experiments show that this feedback enables the controllable robot to exploit space and time that remain unavailable to the conservative baseline. 
The high-dimensional stacking experiments demonstrate that this advantage extends to composite spaces of up to 24 DoF for prescribed task sequences. 
Task dependencies expose the main scalability bottleneck: reactive planning must cover feasible continuations across reachable stochastic branches, while conservative planning can stop after finding one feasible path. 
In the current framework, sparse roadmap coverage can therefore limit complete policy synthesis as the number of modes and branches grows.

Future work will focus on more efficient transition-configuration sampling and vectorized collision checking to improve roadmap coverage and policy synthesis in larger mode graphs.
We will also validate the framework in physics-based simulation environments such as MuJoCo, and on physical hardware.
Further extensions include richer multi-agent skills, learned skill policies, and explicit handling of skill failures such as dropped objects.

{
\footnotesize
\section*{Acknowledgment}
This work was part of an Innovation project supported by Innosuisse.
We acknowledge the use of AI coding agents, including Google Gemini and Claude Code (using Claude Opus and Sonnet 5), to assist in rewriting initial code implementations and writing debugging and test scripts.
}

\bibliographystyle{IEEEtran}
\label{sec:references}
\bibliography{references}

@String(rss = "Proc{.} of Robotics: Science and Systems (R:SS)")

@String(icra = "Proc{.} of the IEEE Int{.} Conf{.} on Robotics and Automation (ICRA)")

@String(iros = "Proc{.} of the IEE/RSJ Int{.} Conf{.} on Intelligent Robots and Systems (IROS)")

@String(icaps = "International Conference on Automated Planning and Scheduling")

@String(tro = "{IEEE} Transactions on Robotics")

@String(ijrr = "International Journal of Robotics Research")

@String(wafr = "World Symposium on the Algorithmic Foundations of Robotics (WAFR)")

@String(ras = "{IEEE} Robotics and Autonomous Systems")

@inbook{Alterovitz2008SMR,
  author = {Alterovitz, Ron and Sim{\'e}on, Thierry and Goldberg, Kenneth},
  title = {The Stochastic Motion Roadmap: A Sampling Framework for Planning with Markov Motion Uncertainty},
  booktitle = rss,
  editor = {Burgard, Wolfram and Brock, Oliver and Stachniss, Cyrill},
  publisher = {The MIT Press},
  address = {Cambridge, MA, USA},
  year = {2008},
  month = {1},
  pages = {233--240},
  isbn = {9780262255868},
  doi = {10.7551/mitpress/7830.003.0031}
}

@misc{Huynh2012IncrementalSampling,
    title={An Incremental Sampling-based Algorithm for Stochastic Optimal Control}, 
    author={Vu Anh Huynh and Sertac Karaman and Emilio Frazzoli},
    year={2012},
    eprint={1202.5544},
    archivePrefix={arXiv},
    primaryClass={cs.RO},
    url={https://arxiv.org/abs/1202.5544}, 
}

@misc{karaman2010incrementalsamplingbasedalgorithmsoptimal,
    title={Incremental Sampling-based Algorithms for Optimal Motion Planning}, 
    author={Sertac Karaman and Emilio Frazzoli},
    year={2010},
    eprint={1005.0416},
    archivePrefix={arXiv},
    primaryClass={cs.RO},
    url={https://arxiv.org/abs/1005.0416}, 
}

@article{Kavraki1996PRM,
  author={Kavraki, L.E. and Svestka, P. and Latombe, J.-C. and Overmars, M.H.},
  journal=tro, 
  title={Probabilistic roadmaps for path planning in high-dimensional configuration spaces}, 
  year={1996},
  volume={12},
  number={4},
  pages={566-580}
}

@misc{Zhao2024iPolicy,
    title={{iPolicy: Incremental Policy Algorithms for Feedback Motion Planning}}, 
    author={Guoxiang Zhao and Devesh K. Jha and Yebin Wang and Minghui Zhu},
    year={2024},
    eprint={2401.02883},
    archivePrefix={arXiv},
    primaryClass={cs.RO},
    url={https://arxiv.org/abs/2401.02883}, 
}

@inproceedings{Hartmann2026SamplingBased,
  title={Sampling-Based Multi-Modal Multi-Robot Multi-Goal Path Planning},
  author={Hartmann, Valentin N and Heinle, Tirza and Huang, Yijiang and Coros, Stelian},
  booktitle = wafr,
  year = {2026},
  date = {2026-06-15},
}

@misc{Mishani2026MOSAIC,
    title={MOSAIC: Skill-Centric Manipulation Planning with Physics Simulation}, 
    author={Itamar Mishani and Yorai Shaoul and Maxim Likhachev},
    year={2026},
    eprint={2504.16738},
    archivePrefix={arXiv},
    primaryClass={cs.RO},
    url={https://arxiv.org/abs/2504.16738}, 
}

@misc{Hedegaard2026TaskSkillPlanning,
    title={Task and Skill Planning: Hierarchical Robot Planning with Black-Box Skills}, 
    author={Benned Hedegaard and Yichen Wei and Ziyi Yang and Ahmed Jaafar and Stefanie Tellex and George Konidaris and Naman Shah},
    year={2026},
    eprint={2504.17901},
    archivePrefix={arXiv},
    primaryClass={cs.RO},
    url={https://arxiv.org/abs/2504.17901}, 
}

@inproceedings{Huang2025APEXMR,
              title = {APEX-MR: Multi-Robot Asynchronous Planning and Execution for Cooperative Assembly},
              author = {Huang, Philip and Liu, Ruixuan and Aggarwal, Shobhit and Liu, Changliu and Li, Jiaoyang},
              year = {2025},
              booktitle = rss,
              url = {https://arxiv.org/abs/2503.15836}
            }

@article{solovey2016drrt,
  author  = {Kiril Solovey and Oren Salzman and Dan Halperin},
  title   = {Finding a needle in an exponential haystack: Discrete RRT for exploration of implicit roadmaps in multi-robot motion planning},
  journal = ijrr,
  volume  = {35},
  number  = {5},
  pages   = {501--513},
  year    = {2016}
}

@article{Chang2026VIMP,
   title={Efficient iterative proximal variational inference motion planning},
   volume={197},
   ISSN={0921-8890},
   url={http://dx.doi.org/10.1016/j.robot.2025.105267},
   DOI={10.1016/j.robot.2025.105267},
   journal=ras,
   publisher={Elsevier BV},
   author={Chang, Zinuo and Yu, Hongzhe and Vela, Patricio and Chen, Yongxin},
   year={2026},
   month=Mar, pages={105267} 
}

@inproceedings{Shah2020AnytimeIntegrated,
  title={Anytime integrated task and motion policies for stochastic environments},
  author={Shah, Naman and Vasudevan, Deepak Kala and Kumar, Kislay and Kamojjhala, Pranav and Srivastava, Siddharth},
  booktitle=icra,
  pages={9285--9291},
  year={2020},
  organization={IEEE}
}

@article{Aghamohammadi2014FIRM,
  title={FIRM: Sampling-based feedback motion-planning under motion uncertainty and imperfect measurements},
  author={Agha-mohammadi, Ali-akbar and Chakravorty, Suman and Amato, Nancy M},
  journal=ijrr,
  volume={33},
  number={2},
  pages={268--304},
  year={2014}
}

@inproceedings{Abbatematteo2024CIP,
  title={Composable interaction primitives: A structured policy class for force-controlled manipulation},
  author={Abbatematteo, Ben and Hedegaard, Benned and Shah, Naman and Konidaris, George},
  booktitle=icra,
  pages={5550--5556},
  year={2024},
  organization={IEEE}
}

@article{Garrett2021TAMP,
  title={Integrated task and motion planning},
  author={Garrett, Caelan Reed and Chitnis, Rohan and Holladay, Rachel and Kim, Beomjoon and Silver, Tom and Kaelbling, Leslie Pack and Lozano-P{\'e}rez, Tom{\'a}s},
  journal={Annual review of control, robotics, and autonomous systems},
  volume={4},
  number={1},
  pages={265--293},
  year={2021},
  publisher={Annual Reviews}
}

@inproceedings{VanDenBerg2005Prioritized,
  title={Prioritized motion planning for multiple robots},
  author={Van Den Berg, Jur and Overmars, Mark},
  booktitle=iros,
  pages={430--435},
  year={2005},
  organization={IEEE}
}

@article{Atzmon2020Robust,
  author={Atzmon, Dor and Stern, Roni and Felner, Ariel and Wagner, Glenn and Bart\'{a}k, Roman and Zhou, Neng-Fa},
  title={Robust Multi-Agent Path Finding and Executing},
  journal={J. Artif. Intell. Res.},
  volume={67},
  pages={549--579},
  year={2020}
}

@inproceedings{Atzmon2020Probabilistic,
  author={Atzmon, Dor and Stern, Roni and Felner, Ariel and Sturtevant, Nathan R. and Koenig, Sven},
  title={Probabilistic Robust Multi-Agent Path Finding},
  booktitle=icaps,
  volume={30},
  pages={29--37},
  year={2020}
}

@inproceedings{Choudhury2020SCoBA,
  author={Choudhury, Shushman and Gupta, Jayesh K. and Kochenderfer, Mykel J. and Sadigh, Dorsa and Bohg, Jeannette},
  title={Dynamic Multi-Robot Task Allocation under Uncertainty and Temporal Constraints},
  booktitle=rss,
  year={2020}
}

@inproceedings{Luna2014BMDP,
  author={Luna, Ryan and Lahijanian, Morteza and Moll, Mark and Kavraki, Lydia E.},
  title={Optimal and Efficient Stochastic Motion Planning in Partially-Known Environments},
  booktitle={AAAI Conf. on Artif. Intell.},
  year={2014}
}

@article{Sutton1999Options,
  author={Sutton, Richard S. and Precup, Doina and Singh, Satinder},
  title={Between {MDPs} and semi-{MDPs}: A framework for temporal abstraction in reinforcement learning},
  journal={Artificial Intelligence},
  volume={112},
  number={1-2},
  pages={181--211},
  year={1999}
}

@inproceedings{Hoenig2016MAPFPOST,
  author={H\"{o}nig, Wolfgang and Kumar, T. K. Satish and Cohen, Liron and Ma, Hang and Xu, Hong and Ayanian, Nora and Koenig, Sven},
  title={Multi-Agent Path Finding with Kinematic Constraints},
  booktitle=icaps,
  year={2016}
}

\end{document}